\documentclass[journal]{IEEEtran}
\usepackage{amsmath,amsfonts}
\usepackage{algorithmic}
\usepackage{algorithm}
\usepackage{array}
\usepackage{textcomp}
\usepackage{url}
\usepackage{verbatim}
\usepackage{graphicx}
\usepackage{cite}
\usepackage{bm}
\usepackage{hyperref}
\usepackage{makecell}
\usepackage{multirow}
\usepackage{multicol}
\usepackage{authblk}
\usepackage[justification=centering]{caption}

\usepackage{marvosym}

\title{Drone-NeRF: Efficient NeRF Based 3D Scene Reconstruction for Large-Scale Drone Survey}

\author{Zhihao Jia$^1$, Bing Wang$^2$, Changhao Chen$^{3*\href{mailto:changhao.chen66@outlook.com}{\textrm{\Letter}}}$\\
$^1$Geography and Environment, University of Southampton, Southampton, SO17 1BJ, UK\\
$^2$Department of Aeronautical and Aviation Engineering, The Hong Kong Polytechnic University\\
$^3$College of Intelligence Science and Technology, National University of Defense Technology\\
{\tt\small zj3g20@soton.ac.uk, bingwang@polyu.edu.hk, changhao.chen66@outlook.com}
}

\let\oldtwocolumn\twocolumn
\renewcommand\twocolumn[1][]{%
    \oldtwocolumn[{#1}{
    \begin{center}
    \captionsetup{hypcap=false, justification=justified,singlelinecheck=false}
            \vspace{-2\baselineskip} % Move image up one line
           \includegraphics[width=\textwidth]{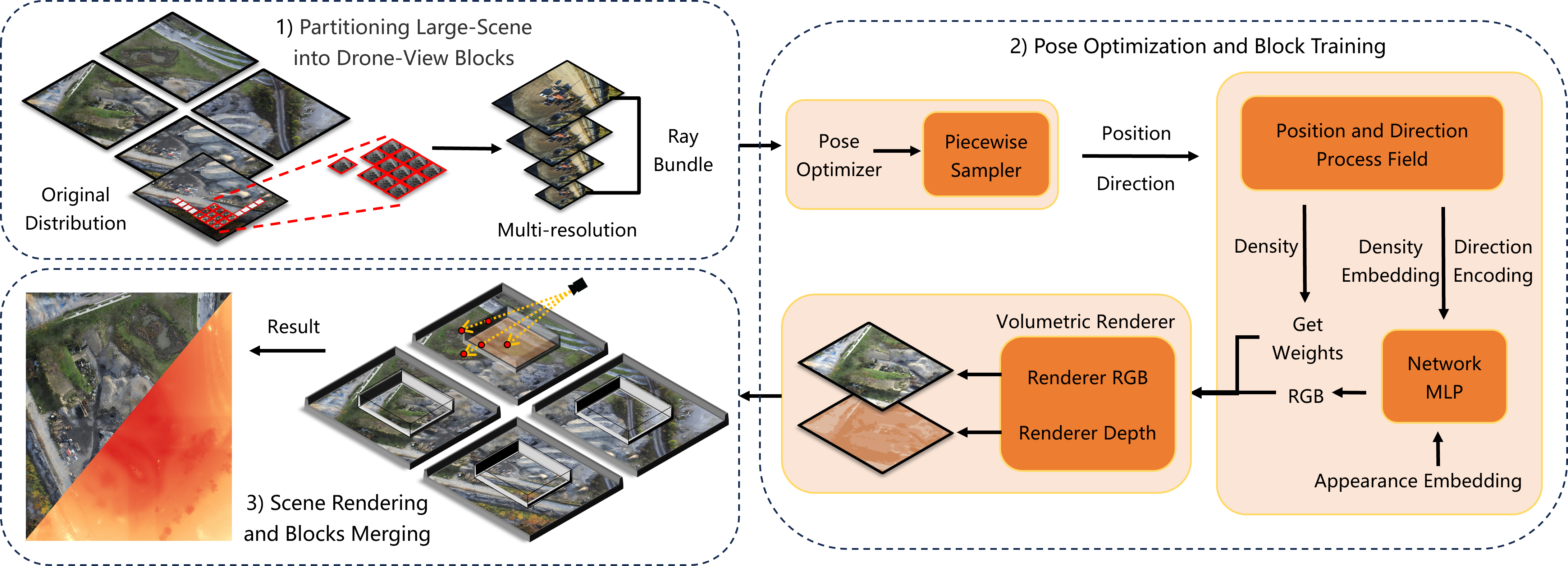}
           \vspace{0em} % Add blank lines and spacing between the legend and picture
           \captionof{figure}{
           The Proposed Drone-NeRF Pipeline. 1) The process involves initial scene segmentation, leading to sub-regions characterized by multi-resolution representations achieved through image down-sampling. 2) Optimization of pose and stratified sampling work to transform spatial data into Euclidean coordinates, thereby guiding the integration of density fields. The fusion of positional and directional data alongside Embedded Appearance cues refines exposure settings, generating Weights and RGB values. This obtains the results for each sub-scene. 3) Volumetric rendering yields confined RGB and Depth attributes while simultaneously addressing shadows originating from vertical planes. The expansion of sub-scene boundaries compensates for occluded regions, resulting in a seamless and refined scene representation through the coherent stages of Drone-NeRF pipeline.
           }
           \vspace{1em} % Add blank lines and spacing between the legend and picture
           \label{fig:Pipeline}
        \end{center}
    }]
}

\begin{document}

\maketitle

\begin{abstract}
Neural rendering has garnered substantial attention owing to its capacity for creating realistic 3D scenes. However, its applicability to extensive scenes remains challenging, with limitations in effectiveness.
In this work, we propose the Drone-NeRF framework to enhance the efficient reconstruction of unbounded large-scale scenes suited for drone oblique photography using Neural Radiance Fields (NeRF). Our approach involves dividing the scene into uniform sub-blocks based on camera position and depth visibility. Sub-scenes are trained in parallel using NeRF, then merged for a complete scene. We refine the model by optimizing camera poses and guiding NeRF with a uniform sampler. Integrating chosen samples enhances accuracy. A hash-coded fusion MLP accelerates density representation, yielding RGB and Depth outputs. Our framework accounts for sub-scene constraints, reduces parallel-training noise, handles shadow occlusion, and merges sub-regions for a polished rendering result. This Drone-NeRF framework demonstrates promising capabilities in addressing challenges related to scene complexity, rendering efficiency, and accuracy in drone-obtained imagery.

% In this work, we propose the Drone-NeRF framework to enhance the efficient reconstruction of unbounded large-scale scenes suited for drone oblique photography using Neural Radiance Fields (NeRF). Our approach involves dividing the scene into uniform sub-blocks based on camera position and depth visibility. This reduces noise and improves clarity. Sub-scenes are trained in parallel using NeRF, then merged for a complete scene. We refine the model by optimizing camera poses and guiding NeRF with a uniform sampler. Integrating chosen samples enhances accuracy. A hash-coded fusion MLP accelerates density representation, yielding RGB and Depth outputs. Our framework accounts for constraints, reduces noise, handles shadow occlusion, and merges sub-regions for a polished result. This Drone-NeRF framework demonstrates promising capabilities in addressing challenges related to scene complexity, rendering efficiency, and accuracy in drone-obtained imagery.
\end{abstract}


\section{Introduction}
\IEEEPARstart{S}{cene} reconstruction stands as a pivotal challenge within the realm of computer vision. In this context, Unmanned Aerial Vehicles (UAVs) have risen to prominence due to their inherent advantages, as they can serve as valuable tools for conducting field surveys and constructing three-dimensional (3D) representations of scenes \cite{inzerillo2018image, zhao2021structural, shang2018real, jiang2020uav}. By leveraging their ability to capture aerial images from a bird's-eye view, drones effectively bypass obstacles encountered at ground level, such as potholes, steps, and water bodies. These aerial images cover a wide range of landscapes and maintain a  level of overlap among consecutive frames. The continuous improvement of sensor technology and complementary drone equipment has notably enhanced the quality of these images. Consequently, this imagery has now reached a level of adequacy that is essential for accomplishing 3D reconstructions of expansive scenes from the top-down perspective that drones offer.

Typical conventional approach to 3D reconstruction includes Structure From Motion (SfM) in conjunction with Multi-View Stereo \cite{snavely2006photo, shen2013accurate}. However, using SfM in larger scenes is problematic due to its time-consuming nature and limited effectiveness. The process of reconstructing large-scale scenes requires a significant amount of resources and time, which hinders the quick achievement of the desired reconstruction results \cite{xu20193d}. Moreover, it's difficult to reconstruct complex scenes with lots of details unless you already know the depths of different points. Additionally, when dealing with scenes that have intricate elements like water bodies or small objects, SfM encounters difficulties \cite{vu2011high}. These challenges arise mainly from not having enough accurate reference points to work with \cite{carr2001reconstruction}. 

%Recent scholarly attention has gravitated towards neural rendering techniques. 
Recently, there have been growing interests in neural rendering technique, i.e. Neural Radiance Fields (NeRF), due to its ability to create highly realistic reconstructions of scenes when given information about camera viewpoints \cite{mildenhall2021nerf}. Compared with traditional SfM, NeRF owns advantages, such as reconstructing more realistic scene, generating new views of scenes from a given pose, and creating 3D reconstructions more efficiently \cite{ost2021neural, song2022implicit, mari2022sat, noonan2021neuralplan}. Though NeRF has shown impressive results, especially for scenarios focused on individual objects, there are still limitations, such as it is hard to generalize to large scenes. Block-NeRF \cite{tancik2022block} solves this problem by dividing a large scene into small blocks. This can be useful for things like vehicle perspectives, but it requires a large amount of computational resources. Also, it often struggles when representing larger and more complex scenes because there's a limit to how much information they can handle, leading to longer reconstruction times as scenes become more complicated. Additionally, the paths drones take when flying are affected by factors like air resistance, which introduces errors in the data from sensors that tell the drone where it is and which way it's pointing.

To optimize the reconstruction of expansive scenes, we propose Drone-NeRF, a novel neural rendering framework that is suitable for drone oblique photography. As demonstrated in Fig.\ref{fig:Pipeline}, initially, the scene is partitioned into uniform sub-blocks, each bounded by the highest camera position and the maximum depth visibility range. 
Based on these sub-scene ranges, it is possible to exclude training scene noise outside the sub-region and redundant scenes outside that region, and it is also possible to eliminate the overlap of these noises with other respective sub-scenes. Therefore, The correct merging of sub-scenes is ensured by sub-scene constraints and noise removal to avoid interference from external scenes and noise, which can improve the rendering quality.
This range-based partitioning reduces noise and enhances scene quality. These sub-blocks are independently trained using neural rendering. Merging and rendering sub-scenes based on their locations generates the complete scene. For model enhancement, a camera optimization model facilitates loss gradient backpropagation to input pose computation. A uniform sampler guides NeRF for preliminary scene insight, enabling efficient fine sampling for refinement. Combining these chosen samples with influential scene facets enhances accuracy. Spatial distortion addresses unbounded scenes. A compact fusion MLP with hash coding represents scene density, aiding swift queries and guided sampling. Positional hashing, coupled with spherical harmonics encoding, yields RGB and Depth outputs. Applying scene constraints and noise reduction to each sub-scene, while addressing shadow occlusion, further refines results. The amalgamation of these sub-regions yields a comprehensive reconstructed area.

% ========================================
In summary, our contributions are as follows.
\begin{itemize}
    \item We propose Drone-NeRF, a novel neural radiance rendering (NeRF) framework that allows efficient 3D drone-view scene reconstruction in large-scale environments.
    \item We present a solution to partition large-scale scene into drone-view blocks and allow effective parallel NeRF training within each block with fast convergence and spatial distortion correction.     
    \item Our approach allows large-scale scene rendering by merging the partitioned drone-view blocks flexibly. In addition, our model can easily expand the sub-bounding box for rendering to mitigate the impact of shadows in the edge area.
\end{itemize}
% ========================================

\section{Related Work}
\subsection{Traditional 3D reconstruction}
The techniques employed in 3D reconstruction can be broadly categorized into active and passive fusion approaches \cite{hao2022review}. The passive method typically utilizes the surrounding environment, such as natural light reflections, in conjunction with images acquired by cameras to compute the three-dimensional spatial information of objects via specific algorithms \cite{xu2022depth}. Notably, the Shape From Shading (SFS) method plays a key role in this category. It determines the orientation of object surface points, i.e., the normal vector direction, by analyzing changes in light and shadow on the surface of the object \cite{horn1986variational}.
Another prominent passive method is Photometric-Stereo, which simplifies the determination of the normal vector and employs three maps captured under different lighting conditions but the same viewpoint, utilizing stereo vision principles \cite{woodham1979photometric}. Additionally, Multi-View Stereo (MVS) significantly relies on parallax or the disparity in the projection position of the same 3D point observed by different cameras, serving as a primary approach for recovering 3D structures \cite{jin2005multi}. Operating on the principle of triangulation, MVS computes the 3D position of a third point by intersecting two rays emitted from two known 3D points \cite{seitz2006comparison}. These approaches contribute significantly to the advancement of 3D reconstruction, enabling accurate and detailed representations of real-world objects and scenes \cite{xi2022raymvsnet}.
%(Please write the limitations of traditional methods for drone survey in large scale environment)
Traditional scene reconstruction methods have some limitations. The reconstruction effect of the water environment is usually poor \cite{carr2001reconstruction}. Secondly, when reconstructing small objects in the scene, it is often impossible to restore them completely \cite{vu2011high}. In addition, scene reconstruction needs to consume a lot of time and hardware resources, and it is difficult to quickly obtain the reconstruction effect \cite{ xu20193d}.

\subsection{Neural Radiance Fields (NeRF)}
Neural Radiance Fields (NeRF) represent a groundbreaking approach that achieves realistic novel viewpoint rendering, effectively capturing scene geometry and view-dependent effects \cite{mildenhall2021nerf}. NeRF leverages the multilayer perceptron (MLP) to project 3D positions and orientations into the radiation field, enabling the training of each scene in this way. However, successful application of the NeRF model necessitates input photos with a certain level of quantity and clarity, while ensuring accurate pose estimation and consistent lighting conditions at a similar scale. Under these specific prerequisites, the NeRF model can synthesize highly detailed novel views \cite{rematas2022urban}. 
the remarkable performance of NeRF has inspired numerous extensions and explorations across diverse domains. Notable extensions include rendering portraits of people \cite{cai2022pix2nerf, athar2022rignerf, krishnan2023novel}, human modelling \cite{xu2022surface, shao2022doublefield, weng2022humannerf}, editable scene geometry and textures \cite{yang2022neumesh, athar2022rignerf, yang2021learning, liu2021editing}, high dynamic range view synthesis \cite{mildenhall2022nerf, huang2022hdr, jun2022hdr}, and realization of reconstruction of large indoor scenes \cite{zhu2022nice, yang2022vox}, among others. These advancements showcase the versatility and potential of NeRF as a foundational framework that continually fuels progress and innovation in the field of computer graphics and computer vision.
% (Please write the limitations of current NeRF methods for drone survey in large scale environment)
NeRF-based methods show better performance on target objects. However, as the scene expands, it takes more hardware resources and time to reconstruct the corresponding scene. And due to limited model capacity, it is usually not possible to reconstruct large scenes.

\subsection{3D Reconstruction in Urban-Scale Environments} 
Urban Radiance Fields, such as CityNeRF, BlockNeRF, and Mega-NeRF represent notable approaches tailored for city-level environments \cite{rematas2022urban, xiangli2021citynerf, tancik2022block, turki2022mega}. Urban Radiance Fields effectively leverages RGB images and LiDAR scan data to facilitate 3D reconstructions and novel view compositions \cite{rematas2022urban}. CityNeRF adopts a progressive learning strategy, which activates high-frequency channels in positional encoding, preserving intricate details during training \cite{xiangli2021citynerf}. BlockNeRF, on the other hand, trains individual small blocks within scene subdivisions, allowing for block-by-block adjustments \cite{tancik2022block}. The model also incorporates exposure conditioning to control camera exposure and utilizes appearance embedding to address the influence of varying weather conditions \cite{MartinBrualla2020NeRFIT}. Furthermore, the model fine-tunes pose estimation and effectively excludes moving objects in the scene, ensuring more accurate representations.
To address the low-resolution rendering issue, the model introduces techniques from mip-NeRF \cite{Barron2021MipNeRFAM}. Mega-NeRF, on the other hand, employs NeRF++ to handle expansive borderless scenes and enhance the background's expressive capacity \cite{Zhang2020NeRFAA}. It also utilizes the appearance embedding method from NeRF in the Wild to balance the influence of varying lighting conditions across different images in the same scene \cite{MartinBrualla2020NeRFIT}. Additionally, the method of PlenOctrees is incorporated to accelerate the rendering process, setting a threshold to halt sampling when the accumulated transmittance exceeds it, thereby further improving rendering speed \cite{Yu2021PlenOctreesFR}. These methods collectively contribute to advancements in city-level scene understanding and rendering, facilitating applications in urban environments.

%\section{Primer}

\section{Method} 
\subsection{Basic NeRF Framework}
% introduce basic NeRF framework here.

We begin by providing an overview of the basic Neural Radiation Field (NeRF) framework \cite{mildenhall2021nerf}. %Consequently, NeRF introduces a pioneering avenue for attaining exceptional 3D rendering image generation.
%NeRF represents a deep neural network-based method for generating top-tier 3D rendering images. 
%NeRF allows 3D rendering image generation.
NeRF employs a multi-layer perceptron (MLP) to model and encode the radiation field within a given scene, thus enabling implicit scene modelling. By training MLP on a collection of images with known camera poses, its objective is to minimize the discrepancy between the predicted and actual pixel colours from various viewing angles. As a result, the trained neural network exhibits the capability to produce good-quality images from any arbitrary viewpoint. 

Specifically, in NeRF, each image pixel is associated with a single camera ray \textbf{\textit{r}}. This ray is then sampled along its trajectory. At each sample point $p_{i}$, the Multi-Layer Perceptron (MLP) is queried with both position and viewing direction information \textbf{\textit{(x, y, z, $\theta$, $\phi$)}}. The MLP provides the opacity \textbf{\textit{$\sigma$}} and color values \textbf{\textit{(r, g, b)}} corresponding to the 3D position. The voxel density is computed using an MLP network, enabling the calculation of the gradient with respect to \textbf{\textit{$\sigma$}} through the backpropagation algorithm. This procedure allows expressing the colour \textsc{C}(\textbf{\textit{r}}) of this camera ray \textbf{\textit{r}}(t) = \textbf{\textit{o}} + t\textbf{\textit{d}} with near and far bounds $t_n$ and $t_f$ using an integral equation denoted by Eq.\ref{C(r)}. The term $\exp\left( {- \int_{t_n}^{t}\sigma\left( {\mathbf{r}(s)} \right)ds} \right)$ represents the cumulative transmittance from $t_n$ to $t$. That is, it is expressed as the probability of not hitting any particle in $t_n$ to $t$.

\begin{gather}
    C(\mathbf{r}) = \int_{t_n}^{t_f} e^{-\int_{t_n}^{t} \sigma(\mathbf{r}(s)) ds} \sigma(\mathbf{r}(t)) \mathbf{c}(\mathbf{r}(t),\mathbf{d}) dt\label{C(r)}
\end{gather}

Since NeRF does not inherently compute continuous 3D point information, using numerical approximation methods becomes necessary. To achieve this, the integration domain along the ray is discretized into \textbf{\textit{N}} segments, and uniform random samples are drawn within each segment. The sample point $i$ is mathematically represented as Eq.\ref{t_{i}}. By leveraging these sample points, the integral equation is transformed into a more manageable summation form, as depicted in Eq.\ref{{C}(r)}, where $\delta_{i}$ denotes the distance between two adjacent sample points surrounding $p_{i}$.

\begin{gather}
    t_{i} \sim U\left\lbrack {t_{n} + \frac{i - 1}{N}\left( {t_{f} - t_{n}} \right),t_{n} + \frac{i}{N}\left( {t_{f} - t_{n}} \right)} \right\rbrack\label{t_{i}} \\
    \hat{C}(r) = {\sum_{i = 1}^{N}e^{-\sum_{j=0}^{i-1} \sigma_j \delta_j}} \cdot \left( {1 - e^{-\sigma_{i} \cdot \delta_{i}}} \right) \cdot \mathbf{c}_{i}\label{{C}(r)}
\end{gather}

Throughout the training process of NeRF, camera rays undergo sampling via a multi-level voxel sampling procedure that leverages both coarse and fine networks. To better capture high-frequency details and enhance training quality, positional encoding is employed. The ultimate loss function is directly defined on the rendered results as the L2 loss, presented as Eq.\ref{L}, and is composed of the sum of the coarse and fine losses. 

\begin{gather}
    \mathcal{L} = \Sigma_{ \mathbf{r} \in \mathcal{R}}\left\lbrack {\left\| {{\hat{C}}_{c}(\mathbf{r}) - C(\mathbf{r})} \right\|_{2}^{2} - \left\| {{\hat{C}}_{f}(\mathbf{r}) - C(\mathbf{r})} \right\|_{2}^{2}} \right\rbrack\label{L}
\end{gather}

However, when deploying NeRF for drone survey tasks in large-scale environments, the original NeRF exhibits certain limitations, including large computation requirement, hard to model certain details of objects within a large scene, and vulnerability to variable illumination and transient occluders. In the subsequent subsections, we present our proposed Drone-NeRF, which further improves NeRF and is specifically designed for aerial view scenes captured by unmanned aerial vehicles. % The enhancements introduced in Drone-NeRF are strategically devised to overcome the aforementioned limitations and bolster the overall performance of NeRF when applied to aerial view scenes.

\subsection{Partitioning Large-Scene into Drone-View Blocks}
To allow efficient neural rendering in large-scale environments, a possible solution is to partition it into blocks in accordance with the scene requirements, that can be trained and tested in parallel. %This is necessary since feeding a large number of training images simultaneously into one deep neural network model would consume excessive GPU memory and prolong the training duration. 
By dividing the scene into blocks, parallel processing becomes conveniently feasible, thereby reducing the training time and catering to the needs of handling vast scenes. Moreover, the details of scene reconstruction can be modelled more effectively within each small block.
Here, we introduce how to partition large scene into blocks with bounding box. Firstly, according to the position information, bounding box can be calculated, which can be represented by its minimum and maximum coordinate values. In Eq.\ref{BoundingBox}, $x_{\min}$, $x_{\max}$, $y_{\min}$, $y_{\max}$, $z_{\min}$, and $z_{\max}$ respectively denote the minimum and maximum coordinate values of the bounding box along the x, y, and z axes.

%\begin{figure}[H]
\begin{figure}
    \centering
    \captionsetup{justification=justified,singlelinecheck=false}
    % \small % Set text size to \small
    % \scriptsize
    \footnotesize 
    \begin{tabular}{m{4.2cm}<{\centering}m{4.3cm}}
         \includegraphics[width=4.0cm,height=3.3cm]{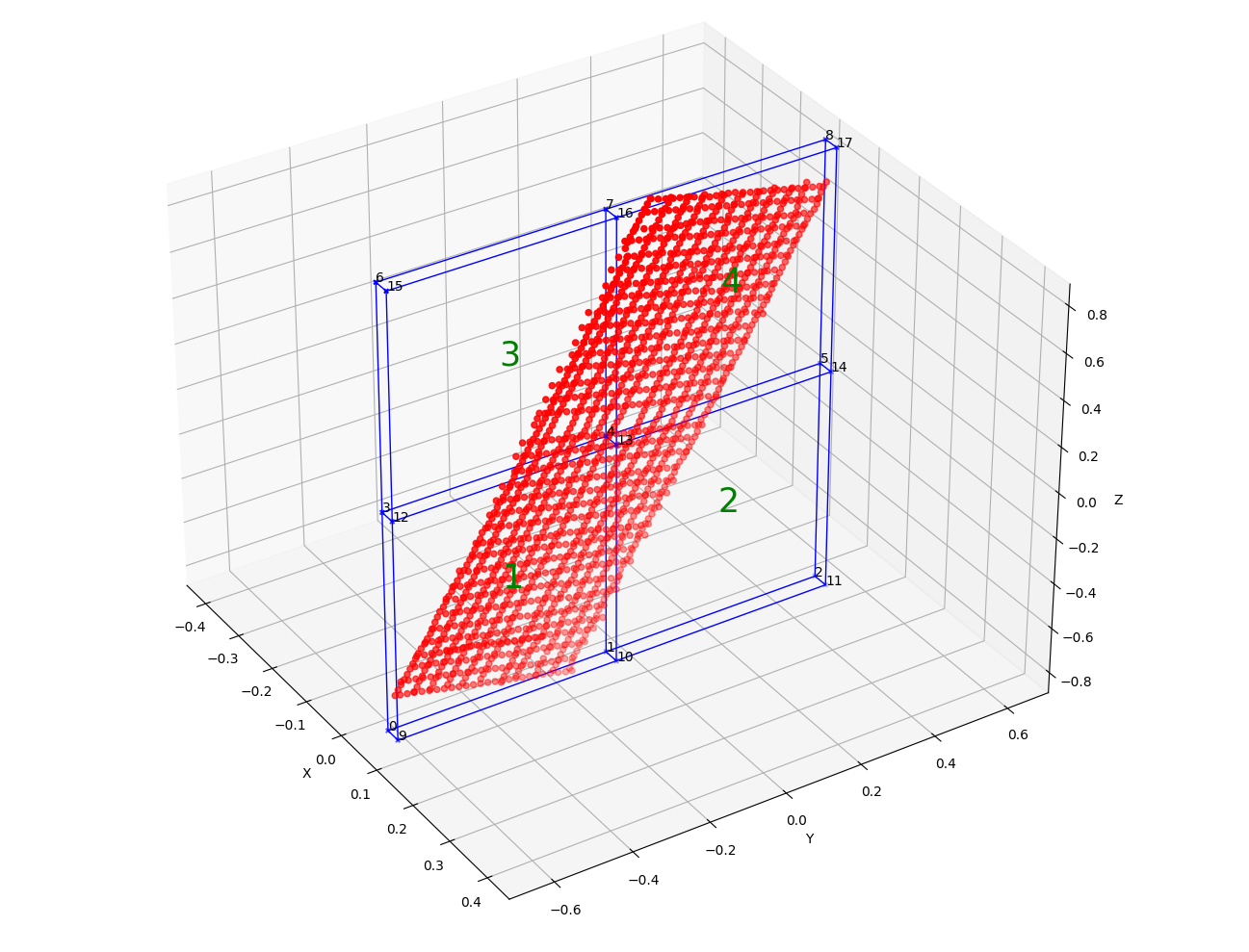}  &  \includegraphics[width=3.85cm,height=3.4cm]{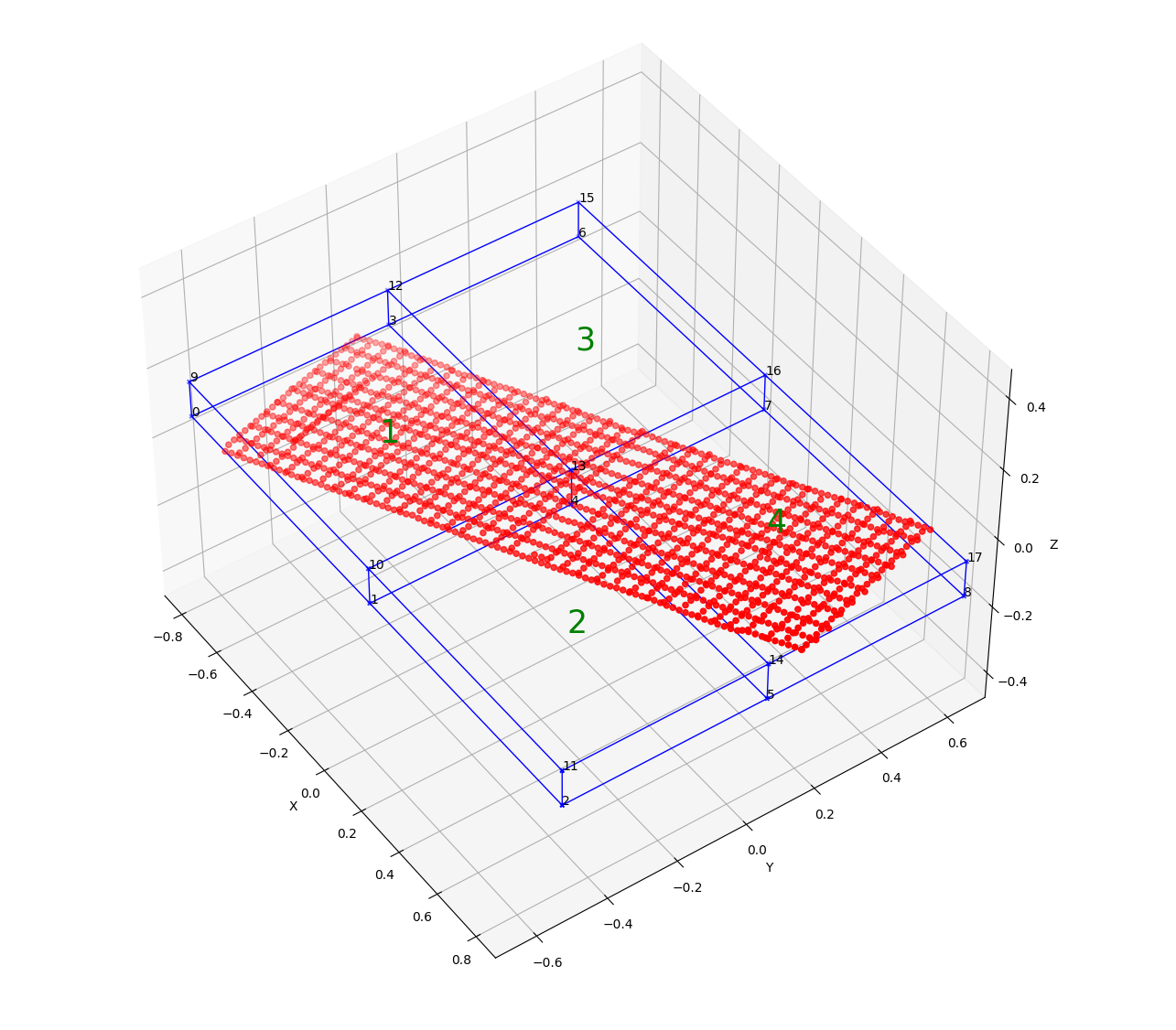} \\

        \multicolumn{1}{c}{\centering {(a) Building poses original}} & \multicolumn{1}{c}{\centering {(b) Building poses transformed}} \\
        
         \includegraphics[width=3.21cm,height=3.4cm]{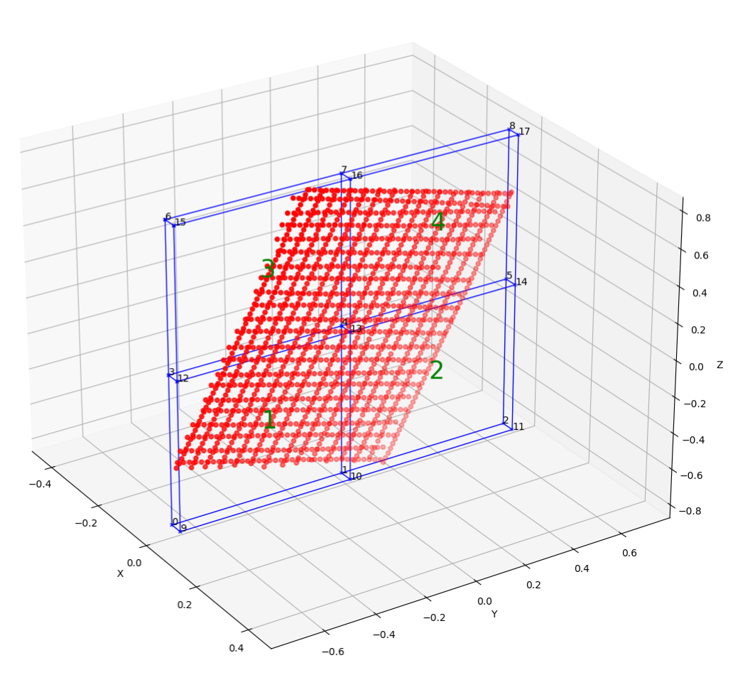} & \includegraphics[width=3.95cm,height=3.4cm]{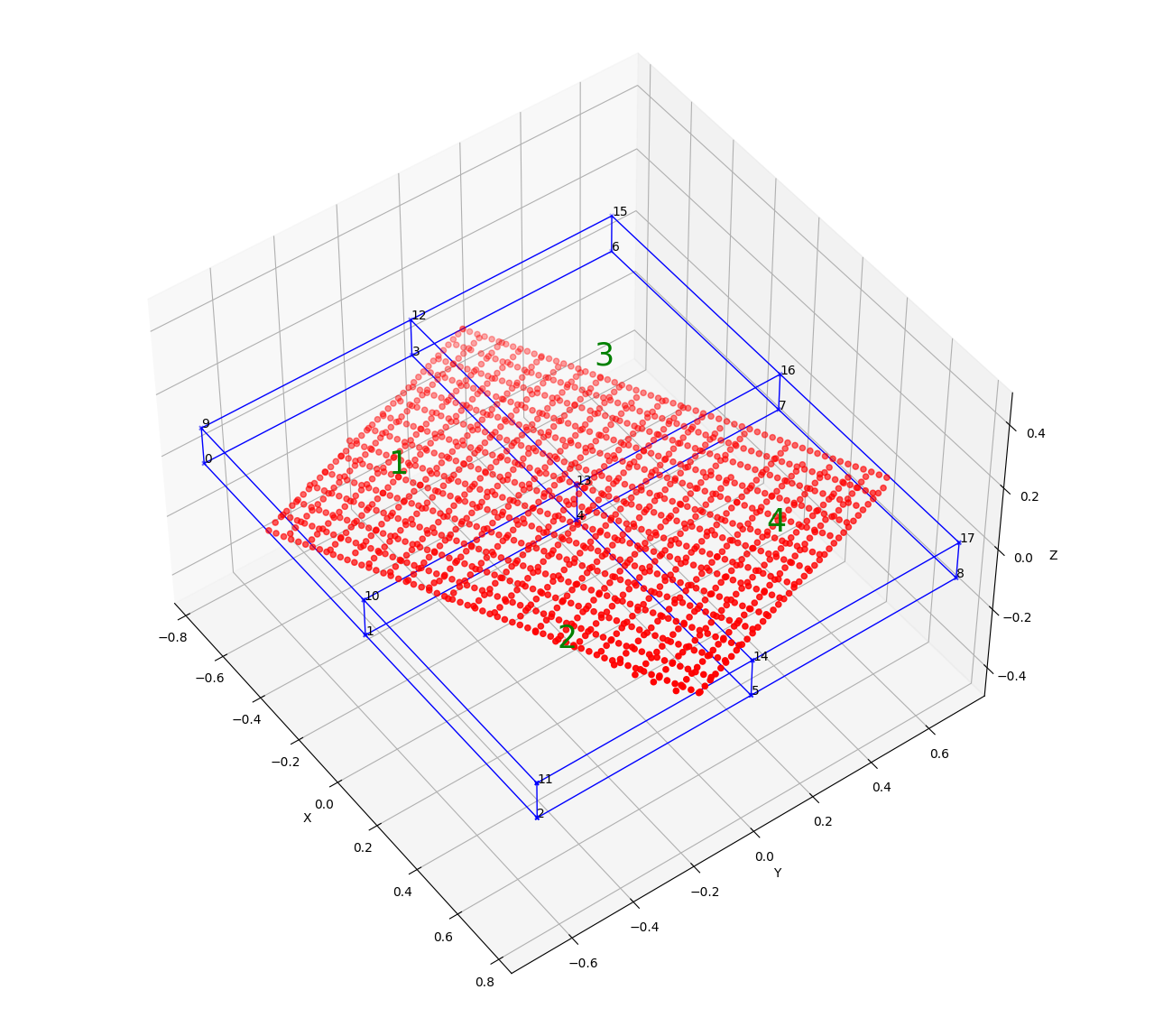}\\
        
        \multicolumn{1}{c}{\centering {(c) Rubble poses original}} & \multicolumn{1}{c}{\centering {(d) Rubble poses transformed}} \\
    \end{tabular}
    \caption{Distribution of image positions in the Building and Rubble Scenes, along with scene subdivision bounding boxes. (a) Image position distribution of the Building Scene (Original). (b) Image position distribution of the Building Scene after rotation transformation (Transformed). (c) Image position distribution of the Rubble Scene (Original). (d) Image position distribution of the Rubble Scene after rotation transformation (Transformed).}
    \label{fig:pos_distribution_and_sub-bounding}
\end{figure}

\begin{gather}
    \mathbf{B} = [\mathbf{x}_{\min}, \mathbf{x}_{\max}] \times [\mathbf{y}_{\min}, \mathbf{y}_{\max}] \times [\mathbf{z}_{\min}, \mathbf{z}_{\max}]\label{BoundingBox}
\end{gather}

The input position is mathematically represented by Eq.\ref{Pos}, while the output is expressed by Eq.\ref{Subboxes}, where each $\mathbf{B}_j$ denotes a sub-bounding box. $j$ is the index of the bounding box. The subdivision of these sub-bounding boxes ensures that the shortest side remains undivided, determined through a comparison of the aspect ratio of the original bounding box. Meanwhile, the remaining sides are divided in accordance with the specified block requirements. Additionally, the sub-bounding boxes possess uniform length and width, with the upper bound of the height being set as the highest position of the image captured by the drone within the block area. The lower bound of the height only needs to meet the scene requirements for enclosing, thereby discarding empty scenes during the subdivision process.

\begin{gather}
    % P = {(x_i, y_i, z_i) \mid i = 1, \ldots, |P|}, \quad n \in \mathbb{N}
    % P={(x_i,y_i,z_i)|i=1,...,|P|}, n\in\mathbb{N} \label{Pos} 
    P = {(\mathbf{x}_i, \mathbf{y}_i, \mathbf{z}_i) \mid i = 1, \ldots, |P|}, \quad n \in \mathbb{N} \label{Pos}
    \\
    \text{Subboxes} ={\mathbf{B}_j \mid j=1, \ldots, n}\label{Subboxes}
\end{gather}

After this, scene transformation is conducted to align the direction of the image with the xy plane consistently, aligning with the perspective of drone aerial photography and facilitating subsequent rendering and merging processes. Different from the Block-NeRF for car-mounted view \cite{tancik2022block}, our work alters the direction of image illumination to match the drone view that enables the NeRF model to capture more relevant information when looking downward while disregarding the sky portion when viewing from above.
Upon transformation, it becomes necessary to adjust the order of the sub-scenes to ensure that the images divided by each sub-scene remain consistent with the state before the transformation. As a result, each sub-scene can be processed individually. The scene position adjustment and the status of the subdivided bounding boxes are illustrated in Fig.\ref{fig:pos_distribution_and_sub-bounding}.

%%%%%%%%%%%%%%%%%%%%%%%%%%%%%%%%%%
% suitable for drone view ()
% Different from Block NeRF (vehicle), 
% Our method can change the direction of image so that in the merging process, our model can focus on real drone view (down, more information), and ignore sky part (up, less information)
%%%%%%%%%%%%%%%%%%%%%%%%%%%%%%%%%%

% \begin{figure}[H]
\begin{figure}
    \centering
    \captionsetup{justification=justified,singlelinecheck=false}
    % \small % Set text size to \small
    % \scriptsize
    \footnotesize 
    \begin{tabular}{m{4.0cm}<{\centering}m{4.0cm}}
        \includegraphics[width=3.5cm,height=2.71cm]{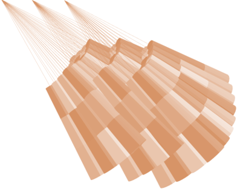}  &  \includegraphics[width=3.5cm,height=2.71cm]{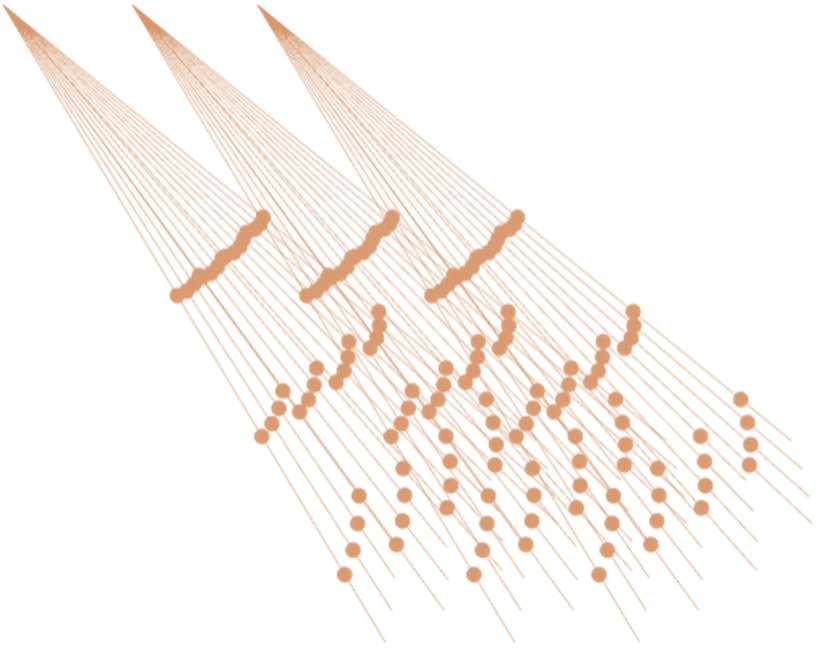} \\

        \thead{ (a) Using conical frustums \\ to represent samples} & \multicolumn{1}{c}{\centering {(b) Representing frustums as points}} \\
         
         \includegraphics[width=4.0cm,height=2.49cm]{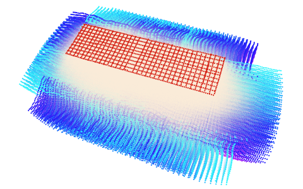}  &  \includegraphics[width=4.0cm,height=2.70cm]{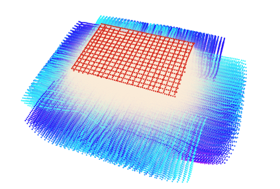} \\

        \thead{(c) Building Scene Coverage} & \thead{(d) Rubble Scene Coverage} \\

    \end{tabular}
    \caption{Illustration of samples in the Building and Rubble Scenes using frustum representations. (a) Samples depicted as conical frustums. (b) Conversion of frustums into points. (c) Coverage of frustum sampling in the Building Scene. (d) Coverage of frustum sampling in the Rubble Scene.}
    \label{fig: conical_frustums}
\end{figure}

\subsection{NeRF Training in Drone-View Block}
Following the partition into sub-bounding boxes, our proposed Drone-NeRF model undertakes parallel training for each sub-scene, i.e. each drone-view block, catering to the specific regions delineated during partitioning. Details of training in each drone-view block can be found below. 

\subsubsection{Image Pyramids}
Prior to NeRF training, the input image undergoes a downsampling operation. By utilizing image pyramids, our Drone-NeRF model can effectively concentrate on the details present at various image levels. It enables enhanced adaptability to diverse spatial scales and contributes to improved performance across the model training process.

\subsubsection{View Frustums}
The camera employed for capturing the image follows a drone-view  model. 
To be suitable to the drone-view images, sampling is carried out on each pixel using frustum, as depicted in Fig. \ref{fig: conical_frustums}, instead of direct ray tracing. This choice is made to achieve anti-aliasing effects and enhance the rendering's capability to represent intricate details. It is important to acknowledge that the pixel area utilized here is an approximation \cite{Barron2021MipNeRFAM}. Our approach effectively balances the trade-off between accuracy and computational efficiency, resulting in high-quality renderings.

\subsubsection{Pose Optimizer}
The ray bundle, formed by the viewing frustum, is directed into the Pose Optimizer. This step facilitates the back-propagation of loss gradients to the input pose calculation, thereby optimizing the pose and enhancing the precision of the rendering process \cite{lin2021barf,tancik2022block}. By iteratively refining the pose parameters, the quality of image rendering is largely improved, resulting in more accurate and visually appealing outputs.

\subsubsection{Multi-Stage Stratified Sampling}
\begin{figure}
    \centering
    \captionsetup{justification=justified,singlelinecheck=false}
    % \small % Set text size to \small
    % \scriptsize
    \footnotesize 
    \begin{tabular}{m{2.3cm}<{\centering}m{4.6cm}}

        \thead{Stage 1: \\ Normal Sampling} & \includegraphics[width=4.6cm,height=1.26cm]{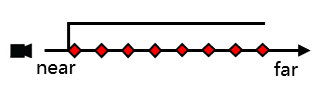} \\
    
        \multicolumn{1}{c}{\centering {}} & \multicolumn{1}{c}{\centering {Convert to position in Euclidean space}} \\
         
        \thead{Stage 2: \\ Stratified Sampling} & \includegraphics[width=4.6cm,height=1.26cm]{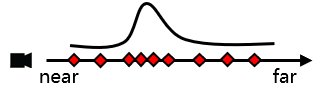} \\

    \end{tabular}
    \caption{ Multi-Stage Sampling. The initial layer comprises regular sampling, uniformly distributing samples across the light. In the second stage, stratified sampling is employed, aligning samples with the probability distribution information.}
    \label{fig: Stratified_sampling}
\end{figure}

Subsequently, segmented sampling is undertaken, commencing with an initial sampling phase where pre-sampling is conducted based on the targeted number of samples. The pre-sampling procedure is implemented using a predefined sampler, and the resulting sampling points are then transformed into positions in Euclidean space. Once the space is sampled, it elucidates which samples significantly contribute to the final color rendering \cite{mildenhall2021nerf}. The second stage involves Probability Distribution Function (PDF) sampling. The sampling density is intensified in regions with larger weights, signifying a higher probability of sampling occurrences in those specific regions \cite{li2022nerfacc}. The two stages employ distinct sampling conditions, as illustrated in Fig.\ref{fig: Stratified_sampling}. This multi-stage sampling approach enables a more efficient and accurate representation of the underlying scene, leading to improved rendering quality and realism.

% \begin{figure}[H]

\subsubsection{Spatial Distortion Correction}
\begin{figure}
    \centering
    \captionsetup{justification=justified, singlelinecheck=false}
    % \small % Set text size to \small
    % \scriptsize
    \footnotesize 
    \begin{tabular}{m{4.0cm}<{\centering}m{4.0cm}}
        \includegraphics[width=4.0cm,height=3.76cm]{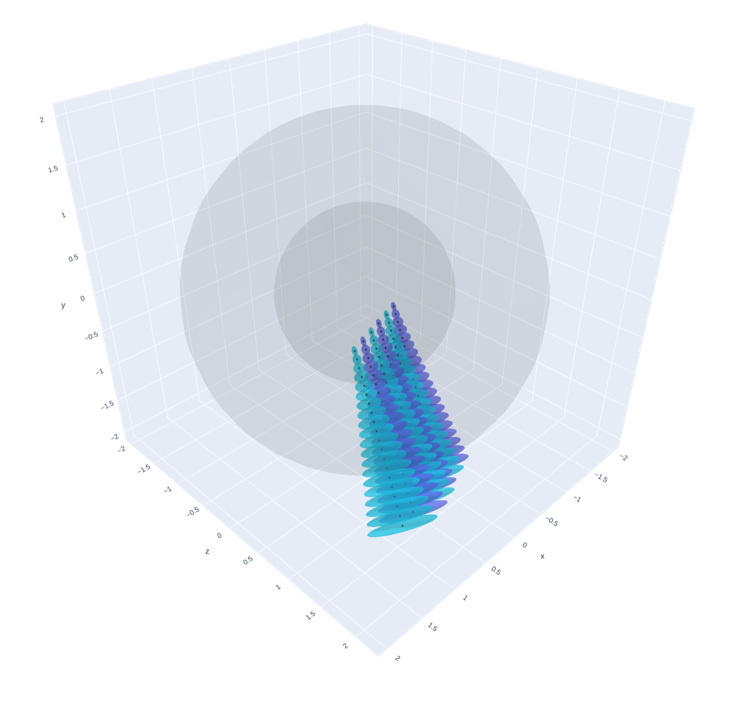}  &  \includegraphics[width=4.0cm,height=3.76cm]{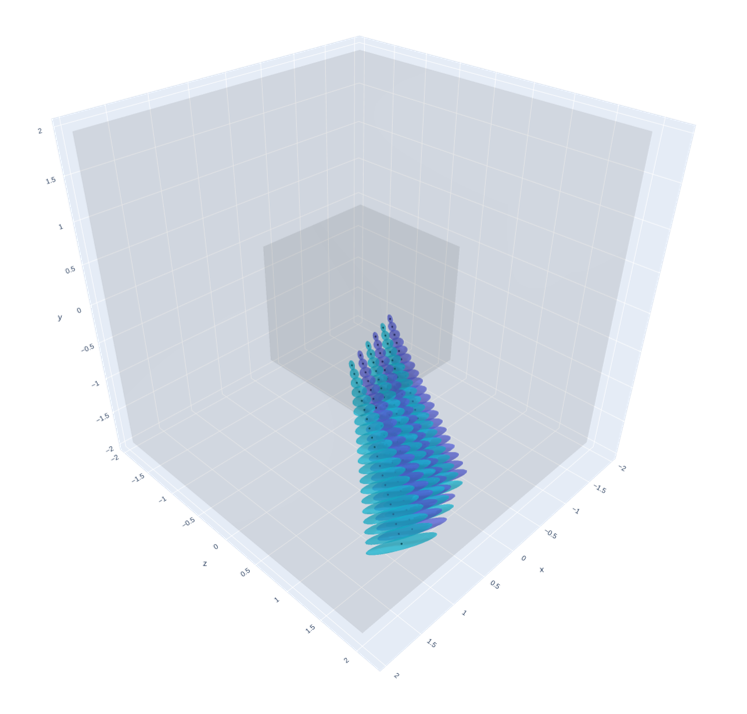} \\
         
         \includegraphics[width=4.0cm,height=3.76cm]{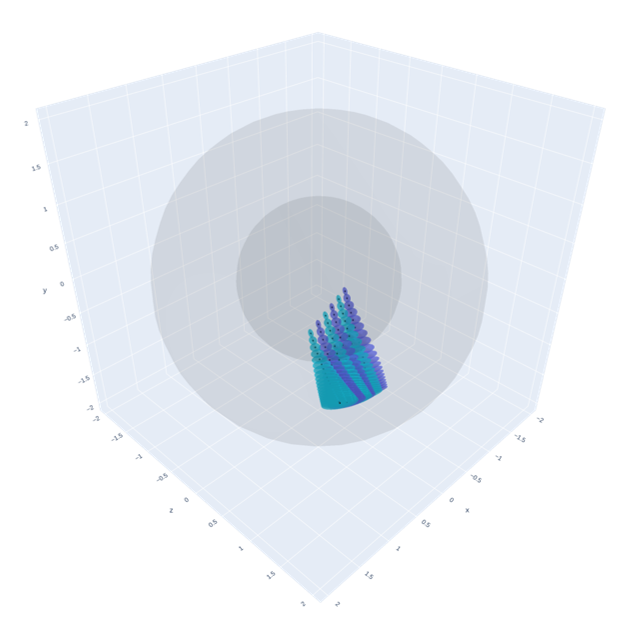}  &  \includegraphics[width=4.0cm,height=3.76cm]{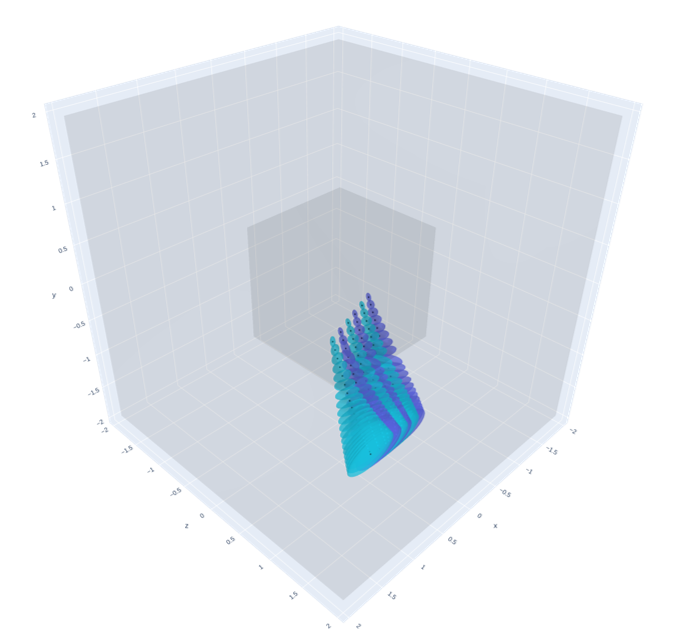} \\

        \multicolumn{1}{c}{\centering {L2 norm scene contraction}} & \multicolumn{1}{c}{\centering {L$\infty$ norm scene contraction}} \\

    \end{tabular}
    \caption{Mapping of unbounded space onto a sphere with a radius of 2 or a cube with a side length of 4. The first row illustrates the state without spatial distortion, while the second row depicts the state with spatial distortion.}
    \label{fig: Contract_unbounded_space_types}
\end{figure}

Upon entering the density field, spatial distortion correction (\ref{Spatial_Distortions}) can be executed to accommodate unbounded scenes effectively. The primary objective of this process is to ensure that scene shrinkage warps unbounded samples into bounded space, thus enabling effective rendering \cite{barron2022mip}. Two types of contraction methods, namely $L_2$ and $L_{\infty}$, are employed, as depicted in Fig.\ref{fig: Contract_unbounded_space_types}. These methods will be thoroughly analyzed in the experimental setup to gauge their effectiveness and impact on the rendering quality and scene representation. $L_{\infty}$ can better represent the cube space scene. Its space contraction can adapt to the geometric state of the sub-scene, and also reduces the excessive distortion of the edge space scene caused by spherical distortion. In addition, different space warp ranges also have an impact on reconstructing the scene. If the scope is not suitable for the scene, the scene within the original reasonable field of view will also be distorted and deformed.

\begin{gather}
\begin{split}
f(x) = \begin{cases} 
    x & \sqrt{x^2} \leq 1 \\
    (2 - \frac{1}{\sqrt{x^2}}) \cdot \frac{x}{\sqrt{x^2}} & \sqrt{x^2} > 1 
\end{cases} 
\end{split}
\label{Spatial_Distortions}
\end{gather}

% \begin{figure}[H]

\subsubsection{Position and Direction Encoding, and Balancing exposure differences}
During the sampling process, the density function plays a critical role in determining the weight of each sample field. This process is designed to adjust the sampling positions, ensuring that the areas in the scene contributing to the final rendering are adequately represented. The density function is realized using a small fused Multi-Layer Perceptron (MLP) along with hash encoding, keeping a balance between accuracy and speed. Consequently, fast queries can be conducted for scene-guided sampling, rendering the overall process efficient and responsive \cite{muller2022instant}.

\begin{figure}
    \centering
    \captionsetup{justification=justified,singlelinecheck=false}
    % \small % Set text size to \small
    % \scriptsize
    \footnotesize 
    \begin{tabular}{m{4.0cm}<{\centering}m{4.0cm}}
        \includegraphics[width=4.0cm,height=3.01cm]{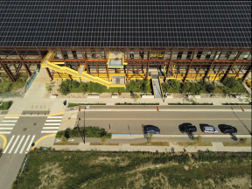}  &  \includegraphics[width=4.0cm,height=3.01cm]{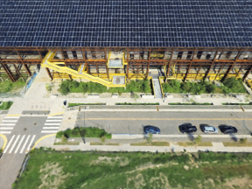} \\

         \thead{Raw image of the Building Scene} & \thead{Image after exposure balance} \\
         
         \includegraphics[width=4.0cm,height=3.01cm]{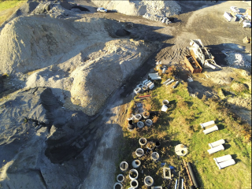}  &  \includegraphics[width=4.0cm,height=3.01cm]{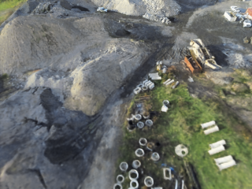} \\

        \thead{Raw image of the Rubble Scene} & \thead{Image after exposure balance} \\

    \end{tabular}
    \caption{Exposure Balancing Effects in Scenes. Within the Building Scene, the initial image exhibits relatively low brightness, which is subsequently increased after exposure balancing, contrasting with the original state. In the Rubble Scene, the initial image demonstrates comparatively high brightness, yet exposure balancing leads to a reduction in brightness compared to the initial image.}
    \label{fig: Appearance_embedding}
\end{figure}

Incorporating Hash Encoding of positions and Spherical Harmonics Encoding of directions \cite{muller2022instant}, the resulting position and direction encoding, along with appearance embeddings \cite{martin2021nerf}, are seamlessly integrated into an MLP. The use of appearance embeddings resolves discrepancies in images arising due to variations in camera exposure, thus achieving the desirable effect illustrated in Fig.\ref{fig: Appearance_embedding}. The MLP subsequently produces the RGB information, culminating in the generation of high-quality renderings that capture intricate scene details and lighting nuances.

% \begin{figure}[H]

%%%%%%%%%%%%%%%%%%%%%%%%%%%%%%
% image pyramid helps model focus on the details of different image levels.

% Fast training

% good for drone large-scale field
%%%%%%%%%%%%%%%%%%%%%%%%%%%%%%

\subsection{Scene Rendering by Merging Essential Drone-View Blocks}
After obtaining the trained model for each sub-scene, employing a bounding box of identical size as the previously divided bounding box to constrain the rendering of the scene effectively removes the noises outside the non-bounding box range. Our Drone-NeRF model facilitates the reconstruction of complete sub-scenes while mitigating interference from excessive noise above the camera's shooting position. Consequently, rendering scenes from a drone's overhead view appear clearer and exhibit improved visual fidelity.

We determine the number of sub-bounding boxes required for merging the rendered scene, according to the intersections between the rays of the rendering area and the sub-bounding boxes. By utilizing only the essential sub-bounding boxes for rendering, the computation requirement for rendering is correspondingly reduced. In instances where the rendering area is confined to a single sub-bounding box, solely that specific box will be utilized for rendering purposes. This selective process optimizes the rendering process and enhances computational efficiency while maintaining the rendering accuracy and quality.

The presence of occlusions caused by the four vertical faces of the sub-bounding box leads to the formation of shadows. These visible shadows result from light hitting the vertical faces and creating projections. To address the influence of these shadows during the merging process, it becomes necessary to introduce a larger bounding box compared to the previous one. In order to mitigate the impact of shadows in the edge area, an expanded sub-bounding box is created. This enlarged sub-bounding box exclusively captures the RGB values corresponding to the projection areas generated by the vertical surface ray irradiation from the smaller bounding box. By utilizing these RGB values in the projection areas, the shadow occlusion regions are effectively replaced, thereby eliminating the influence of shadows. After the area is enlarged, it will overlap with each other. Therefore, the scene colour can be uniform according to the overlapping area. Subsequently, each sub-scene undergoes the same operation, leading to drone-view block merging. Finally, our Drone-NeRF enables the rendering of the overhead view of the drone, free from the artefacts of occlusions and shadows. An overview of our pipeline is shown in Fig.\ref{fig:Pipeline}.

\begin{figure*}
    \centering
    \captionsetup{justification=justified,singlelinecheck=false}
    % \small % Set text size to \small
    % \scriptsize
    \footnotesize 
    \begin{tabular}{m{6.6cm}<{\centering}m{6.6cm}<{\centering}m{6.6cm}}
         \includegraphics[width=6.59cm,height=3.83cm]{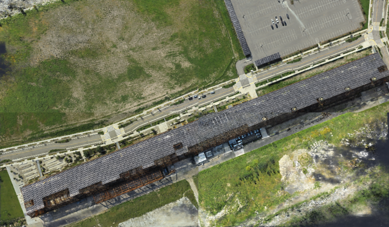}  &  \includegraphics[width=6.59cm,height=3.83cm]{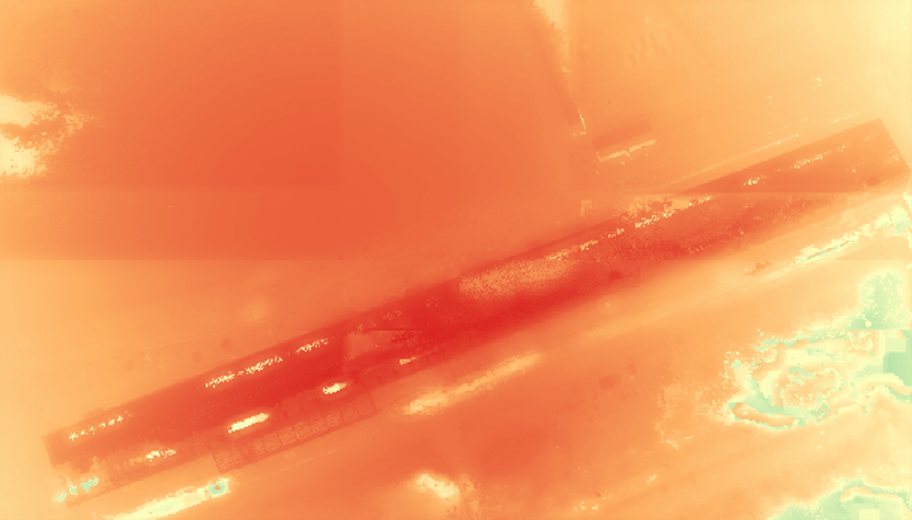} \\

        \multicolumn{1}{c}{\centering {Building Scene}} & \multicolumn{1}{c}{\centering {Building Depth Scene}} \\
        \\
         \includegraphics[width=6.59cm,height=5.39cm]{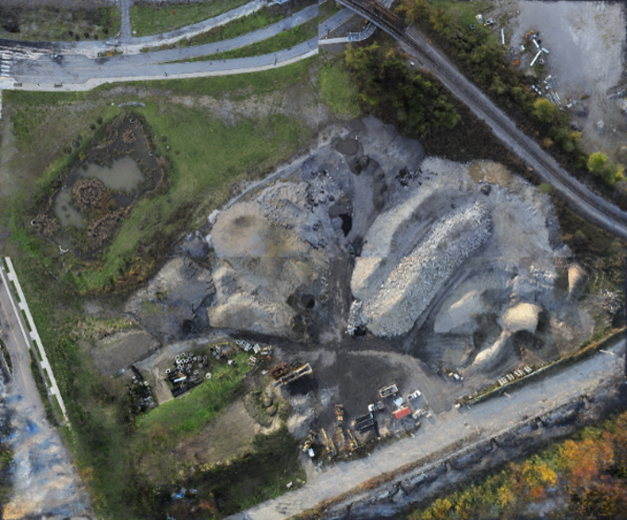} & \includegraphics[width=6.59cm,height=5.39cm]{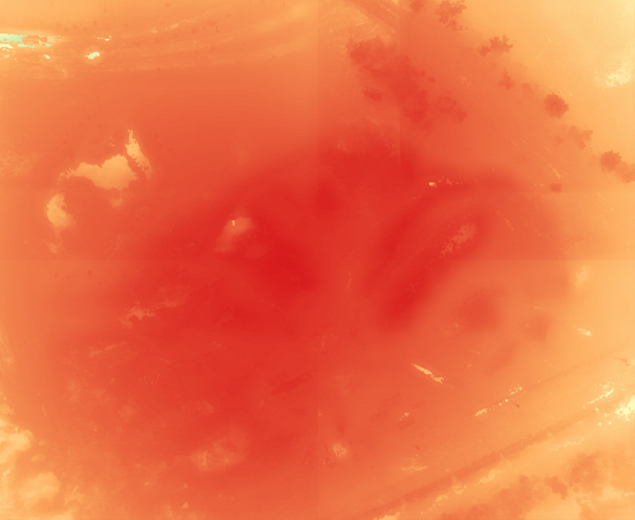}\\
        
        \multicolumn{1}{c}{\centering {Rubble Scene}} & \multicolumn{1}{c}{\centering {Rubble Depth Scene}} \\
    \end{tabular}
    \caption{Scene reconstruction results of 30,000 iterations from the Building Scene and Rubble Scene. The first column showcases the rendering outcomes for the two scenes, while the second column presents the corresponding depth maps. }
    \label{fig:final_result}
\end{figure*}

During the merging process for depth rendering, the depths falling within the scope of the previous larger sub-bounding box are employed for depth merging. Given the overlapping nature of the sub-bounding boxes, only the depths pertaining to each sub-bounding box within the overlapping region require averaging. By comparing the depth value averages across the same overlapping regions, the overall depth of each sub-bounding box is suitably adjusted. This adaptive adjustment process yields smoother and more comprehensive depth information following the merging step. The resulting depth maps exhibit enhanced continuity and coherence, enhancing the overall quality of the merged rendering. The final rendering results and depth maps of the Building and Rubble Scenes are shown in Fig.\ref{fig:final_result}.

%%%%%%%%%%%%%%%%%%%%%%%%%%%%%%%%%%
% Small block (45 degree, drone view)

% large block (remove shade)

% volume density (determine which blocks)
%%%%%%%%%%%%%%%%%%%%%%%%%%%%%%%%%%

\section{Experiments}
%introduce experiments 
In this section, we evaluate the performance of our proposed Drone-NeRF model in two diverse scenarios of public datasets. Additionally, we assess the training efficiency across various models and conduct ablation study over certain parameters, e.g. the frequency of position encoding and spatial warping.

\subsection{Datasets and Training}
\subsubsection{Datasets} 
We assess the performance of our model using two scenarios extracted from the Mill 19 dataset, namely Mill 19-Building (Building) and Mill 19-Rubble (Rubble) \cite{turki2022mega}. In the Building Scene, the dataset comprises footage captured in a grid pattern, covering a substantial area of 500 × 250 ${m^2}$ around an industrial building. As for the Rubble Scene, it encompasses a nearby construction area containing significant debris. To optimize the initial camera poses obtained via GPS in the dataset, we utilize the photogrammetry software \href{https://www.bentley.com/legal/web-properties-terms-of-use/}{ContextCapture}. Through global adjustment optimization, we derive accurate camera pose information.
Upon acquiring camera poses for all images within both scenes of the dataset, the scene bounding box is effectively subdivided. The Building and Rubble Scenes are divided into four sub-blocks. A visual representation of the image positions' distribution and the scene subdivision status can be found in Fig.\ref{fig:pos_distribution_and_sub-bounding}. This rigorous evaluation approach allows us to thoroughly examine the performance and capabilities of our model in realistic and complex real-world scenarios.

\subsubsection{Training} 
The training process adopts a block parallel training process, wherein each sub-block is trained independently, and our model incorporates optimization techniques. Our model consists of a fully connected ReLU layer with 8 layers of 256 hidden units, followed by a final 2 layers of 128 channels. For position encoding, the highest frequency is set to 64, while for direction encoding, it is set to 4. By back-propagating the loss gradient into the input pose computation \cite{lin2021barf}, we obtain valuable information for pose optimization and refinement. This information is then utilized to further optimize and adjust the camera pose, with the camera-optimized learning rate set to 0.0005.

To enhance training efficiency, we employ hierarchical sampling with 128 coarse samples and 256 fine samples per ray, sampling 1024 rays in each batch. The Adam optimizer is utilized with an initial learning rate of 0.005 \cite{kingma2014adam}. Since our model quickly fits the scene, when the number of iterations is too high, the scene will be over-fitted. Therefore all blocks undergo training for 5000 and 30,000 iterations. % It's important to find the right balance to avoid this issue.% 
During the stratified sampling process, the density field is utilized twice. In the initial stage, high-frequency information is not required, so we leverage hash coding in conjunction with a small fusion MLP to facilitate rapid query scenarios \cite{muller2022instant}. This streamlined approach optimizes the training process and contributes to the model's overall efficiency and effectiveness.

\subsubsection{Implementation Details and Baselines} 
In an equivalent training environment, characterized by the utilization of Ubuntu 20.04 as the underlying operating system, an Nvidia 4070ti GPU boasting 12GB of VRAM, a 13th Generation Intel(R) Core(TM) i5-13600KF CPU, and a substantial 32GB memory capacity, our investigation proceeds with meticulous examinations across each distinct sub-block. This encompasses a juxtaposition of two distinct model frameworks, namely MipNeRF and Instant-NGP. Moreover, a comprehensive array of tests is executed, aiming to meticulously assess the extent of spatial distortion within the experimental context. By upholding uniformity in both input content and iterations of training, these thoroughgoing comparisons and evaluations afford the opportunity for a robust appraisal of the model's performance, thus facilitating the derivation of well-founded conclusions grounded in objective observations.

\begin{table*} %table environment, replace [] with h! The effect is the same
\centering % Indicates centered
\captionsetup{justification=justified,singlelinecheck=false}
\renewcommand{\arraystretch}{1.2}
\caption{Comparative Performance Analysis of Various Methods for the Building Scene. Each model undergoes iterations of 5,000 and 30,000 within individual sub-scenes.} 
    \label{tab:building_eval_table}
    \footnotesize 
    \begin{tabular}{c c c c c c c c}
    \hline
       \multirow{2}{*}{\textbf{Methods}} & \textbf{Number of} & \multirow{2}{*}{\textbf{Metric}} & \textbf{Block\_1} & \textbf{Block\_2} & \textbf{Block\_3} & \textbf{Block\_4} & \textbf{Average} 
       \\
       ~ & \textbf{iterations} & ~ & \textbf{793 imgs} & \textbf{192 imgs} & \textbf{192 imgs} & \textbf{763 imgs} & \textbf{tot. 1940 imgs} 
       \\
    \hline
& & PSNR \scriptsize{$\uparrow$} & 17.741 & 16.063 & 18.962 & 16.839 & \textbf{17.401}  \\
Drone-NeRF (Ours) & 5k & SSIM \scriptsize{$\uparrow$} & 0.461 & 0.326 & 0.398 & 0.386 & \textbf{0.392} \\
& & LPIPS \scriptsize{$\downarrow$} & 0.590 & 0.621 & 0.580 & 0.628 & \textbf{0.605}   \\
\hline       
& & PSNR \scriptsize{$\uparrow$} & 17.053 & 15.525 & 17.779 & 16.190 & 16.637  \\
MipNeRF & 5k & SSIM \scriptsize{$\uparrow$} & 0.406 & 0.308 & 0.279 & 0.334 & 0.332  \\
& & LPIPS \scriptsize{$\downarrow$} & 0.723 & 0.764 & 0.754 & 0.747 & 0.747   \\
\hline       
& & PSNR \scriptsize{$\uparrow$} & 17.546 & 16.041 & 18.449 & 16.681 & 17.179  \\
Instant-NGP & 5k & SSIM \scriptsize{$\uparrow$} & 0.418 & 0.351 & 0.309 & 0.338 & 0.354  \\
& & LPIPS \scriptsize{$\downarrow$} & 0.703 & 0.740 & 0.738 & 0.768 & 0.737   \\
\hline
\hline
& & PSNR \scriptsize{$\uparrow$} & 19.002 & 16.085 & 19.997 & 18.775 & \textbf{18.464}  \\
Drone-NeRF (Ours) & 30k & SSIM \scriptsize{$\uparrow$} & 0.550 & 0.439 & 0.500 & 0.470 & \textbf{0.490} \\
& & LPIPS \scriptsize{$\downarrow$} & 0.434 & 0.497 & 0.441 & 0.502 & \textbf{0.469}   \\
\hline       
& & PSNR \scriptsize{$\uparrow$} & 18.970 & 16.641 & 19.255 & 18.077 & 18.236  \\
MipNeRF & 30k & SSIM \scriptsize{$\uparrow$} & 0.426 & 0.360 & 0.309 & 0.351 & 0.361  \\
& & LPIPS \scriptsize{$\downarrow$} & 0.671 & 0.694 & 0.711 & 0.698 & 0.694   \\
\hline       
& & PSNR \scriptsize{$\uparrow$} & 18.037 & 15.184 & 18.744 & 17.715 & 17.420  \\
Instant-NGP & 30k & SSIM \scriptsize{$\uparrow$} & 0.440 & 0.267 & 0.241 & 0.256 & 0.301  \\
& & LPIPS \scriptsize{$\downarrow$} & 0.686 & 0.641 & 0.551 & 0.629 & 0.627   \\
    \hline
   \end{tabular}
\end{table*}

\begin{table*}
\captionsetup{justification=justified,singlelinecheck=false}
    \caption{Comparative Evaluation of Methods within the Building Sub-Scene over 5,000 Iterations. The first column presents the names of the segmented scenes. The second column features the Ground Truth reference. The subsequent three columns offer a comparison of outcomes derived from three distinct models. Highlighted orange and yellow boxes indicate the specific targets chosen for comparison within the Building Scene.}
    \label{tab:building_images_method_comparison_5k}
    \centering
    \renewcommand{\arraystretch}{1.5}
    % \small % Set text size to \small
    % \scriptsize
    {\footnotesize 
    \begin{tabular}{m{1cm}<{\centering}m{3.1cm}<{\centering}m{3.1cm}<{\centering}m{3.1cm}<{\centering}m{3.1cm}<{\centering}m{3.1cm}}
        \hline
        \textbf{Scene} & \textbf{Building} & \multicolumn{3}{c}{\textbf{Methods}} \\
        \cline{3-5}
        \textbf{Block2*2} & \textbf{Ground Truth} & \textbf{Drone-NeRF (Ours)} & \textbf{MipNeRF} & \textbf{Instant-NGP} \\
        \hline
        \\[-3.0ex]
        \multicolumn{1}{c}{\centering {Block\_1}} &  \includegraphics[width=3.06cm,height=2.3cm]{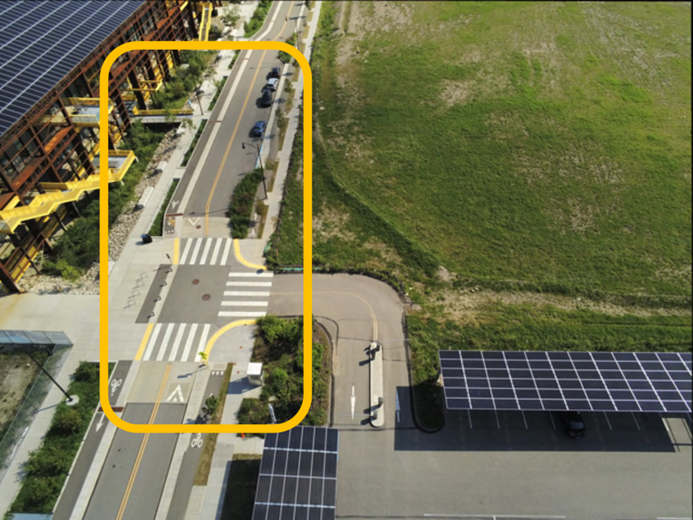}  &  \includegraphics[width=3.06cm,height=2.3cm]{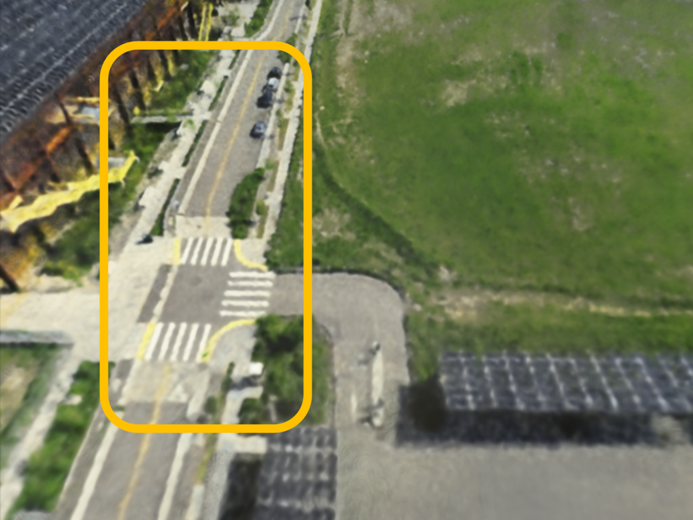}  & \includegraphics[width=3.06cm,height=2.3cm]{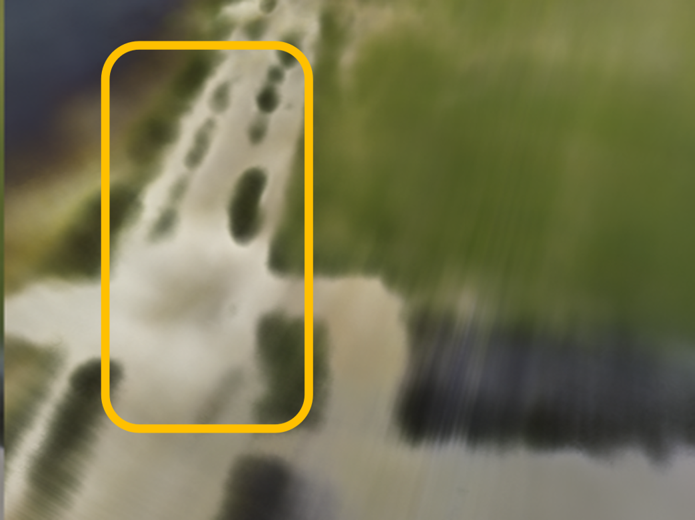}  & \includegraphics[width=3.06cm,height=2.3cm]{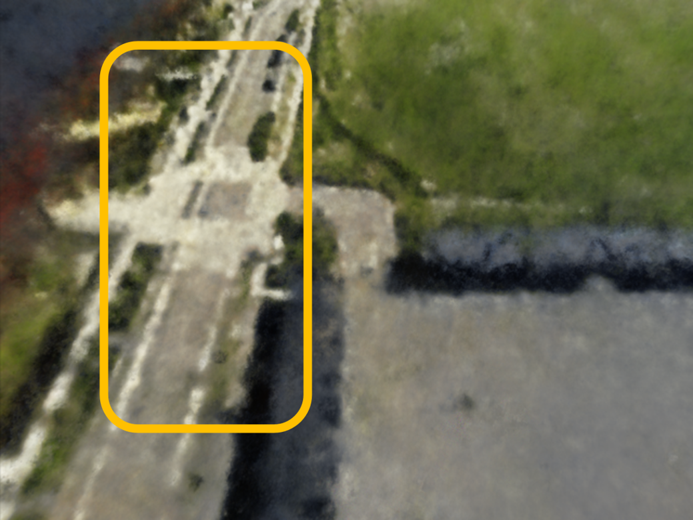} \\
        
        \multicolumn{1}{c}{\centering {Block\_2}} & \includegraphics[width=3.06cm,height=2.3cm]{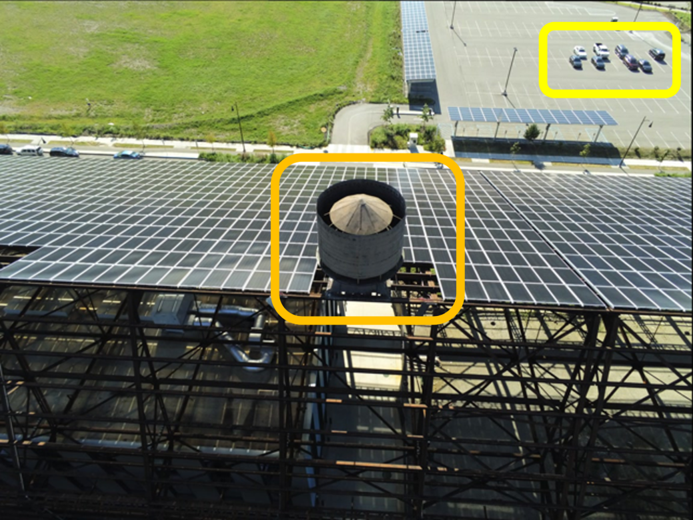} & \includegraphics[width=3.06cm,height=2.3cm]{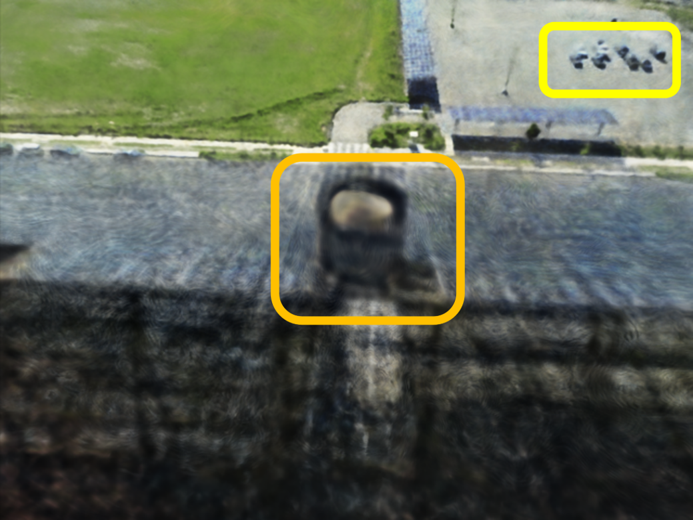} & \includegraphics[width=3.06cm,height=2.3cm]{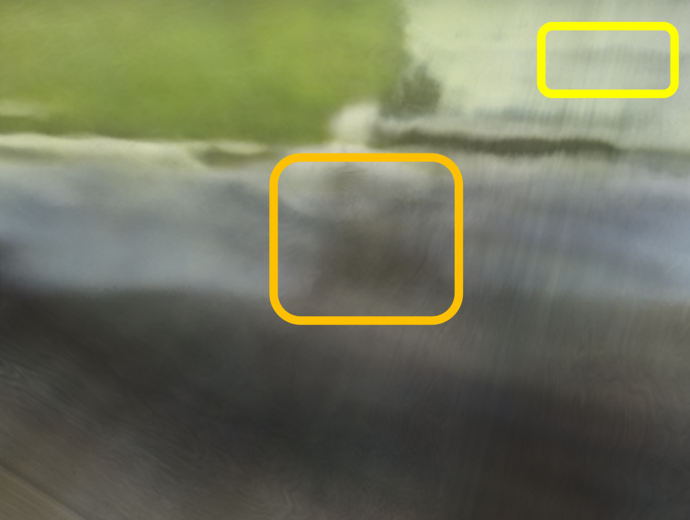} & \includegraphics[width=3.06cm,height=2.3cm]{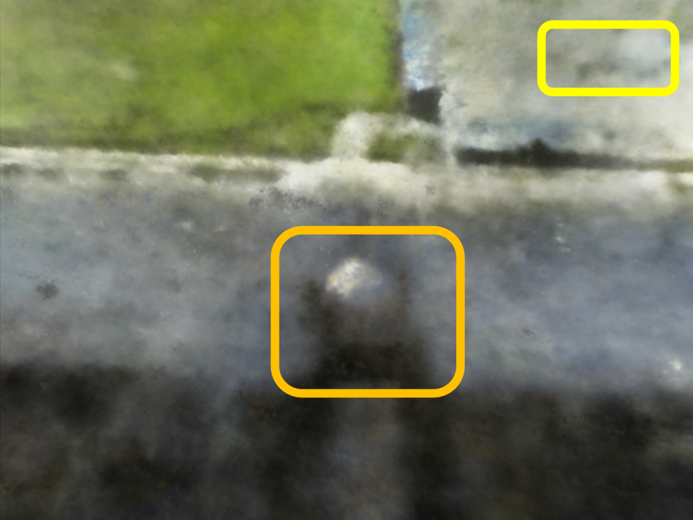}\\
        
        \multicolumn{1}{c}{\centering {Block\_3}} & \includegraphics[width=3.06cm,height=2.3cm]{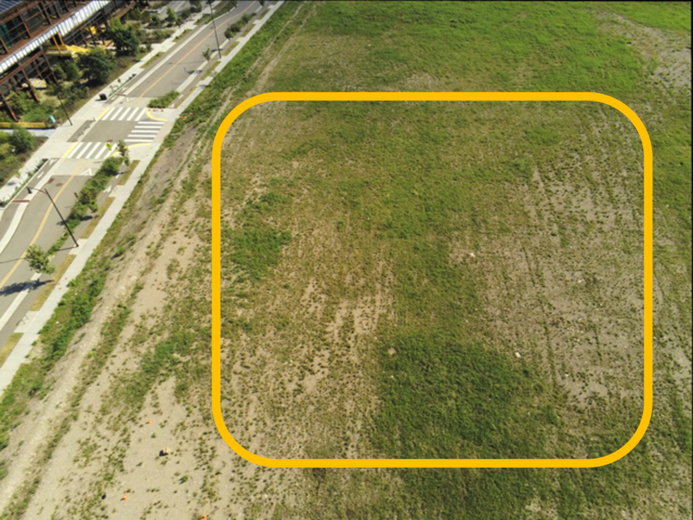} & \includegraphics[width=3.06cm,height=2.3cm]{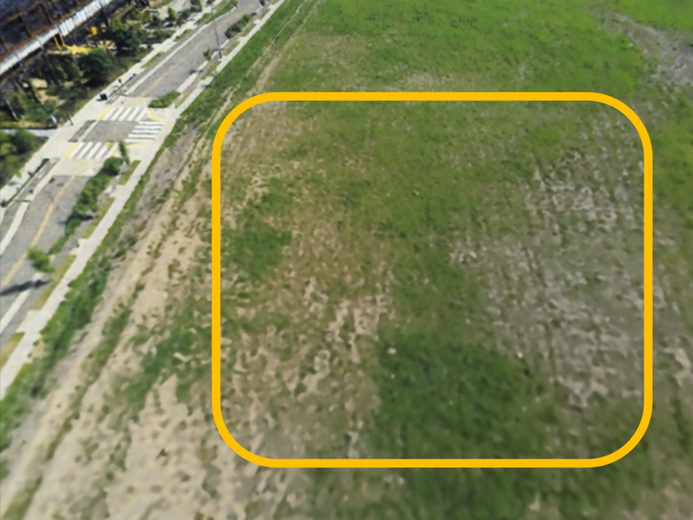} & \includegraphics[width=3.06cm,height=2.3cm]{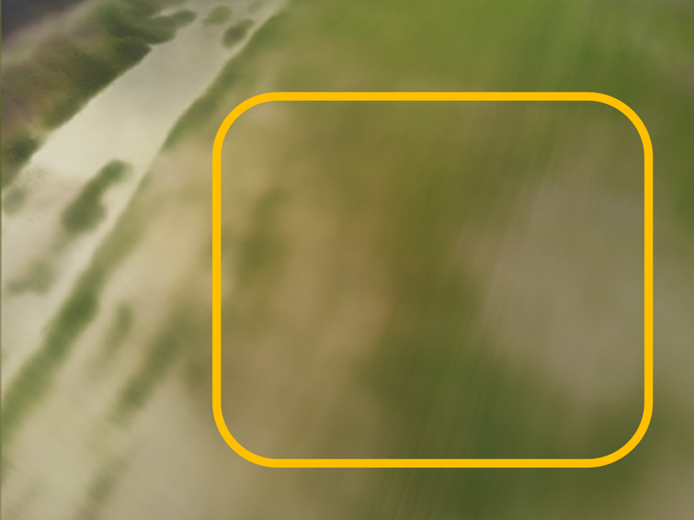} & \includegraphics[width=3.06cm,height=2.3cm]{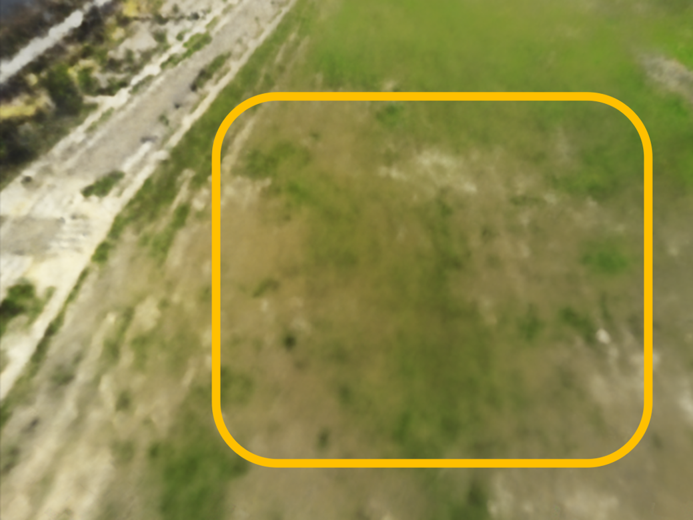} \\
        
        \multicolumn{1}{c}{\centering {Block\_4}} & \includegraphics[width=3.06cm,height=2.3cm]{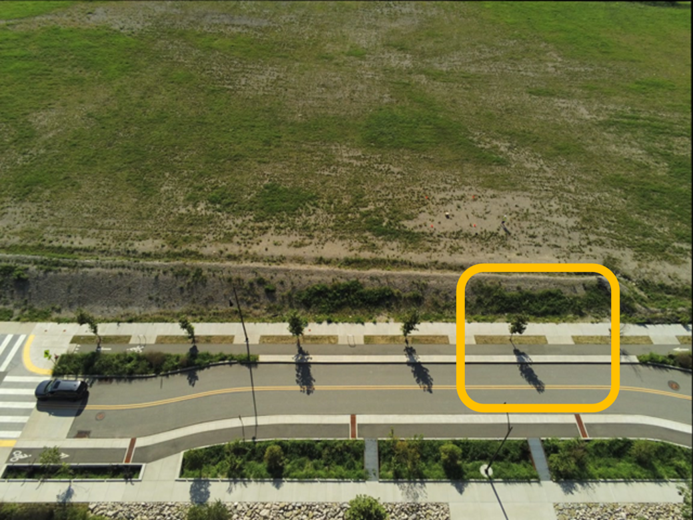} & \includegraphics[width=3.06cm,height=2.3cm]{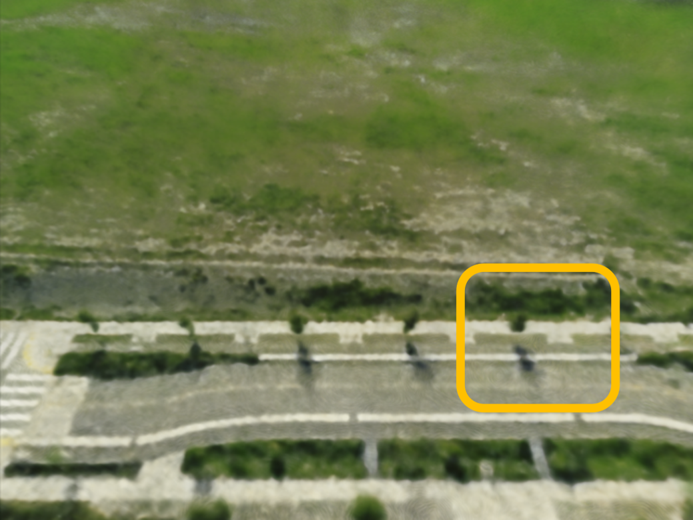} & \includegraphics[width=3.06cm,height=2.3cm]{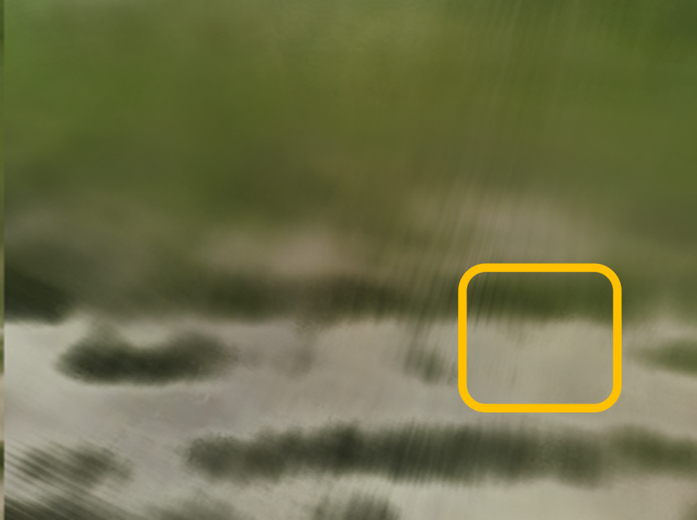} & \includegraphics[width=3.06cm,height=2.3cm]{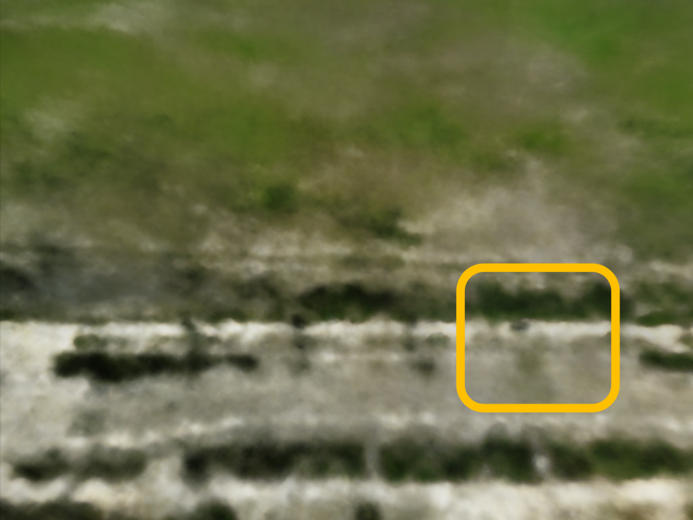} \\
        \hline
    \end{tabular}
    }
\end{table*}

\begin{table*}
    \captionsetup{justification=justified,singlelinecheck=false}
    \caption{Comparative Analysis of Methods within the Building Sub-Scene over 30,000 Iterations. The first column displays the names of the segmented scenes. The second column denotes the Ground Truth reference. Subsequent three columns present a comparison of outcomes obtained from three distinct models. Highlighted orange and yellow boxes identify the specific targets chosen for comparison within the Building Scene.}
    \label{tab:building_images_method_comparison_30k}
    \centering
    \renewcommand{\arraystretch}{1.5}
    % \small % Set text size to \small
    % \scriptsize
    {\footnotesize 
    \begin{tabular}{m{1cm}<{\centering}m{3.1cm}<{\centering}m{3.1cm}<{\centering}m{3.1cm}<{\centering}m{3.1cm}<{\centering}m{3.1cm}}
        \hline
        \textbf{Scene} & \textbf{Building} & \multicolumn{3}{c}{\textbf{Methods}} \\
        \cline{3-5}
        \textbf{Block2*2} & \textbf{Ground Truth} & \textbf{Drone-NeRF (Ours)} & \textbf{MipNeRF} & \textbf{Instance-ngp} \\
        \hline
        \\[-3.0ex]
        \multicolumn{1}{c}{\centering {Block\_1}} &  \includegraphics[width=3.06cm,height=2.3cm]{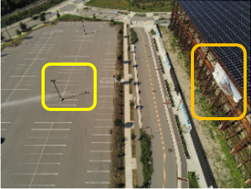}  &  \includegraphics[width=3.06cm,height=2.3cm]{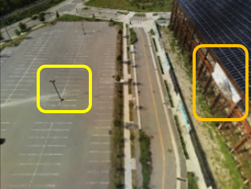}  & \includegraphics[width=3.06cm,height=2.3cm]{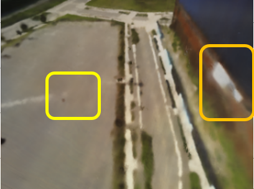}  & \includegraphics[width=3.06cm,height=2.3cm]{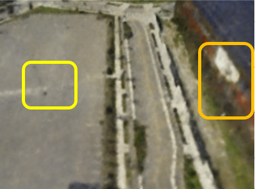} \\
        
        \multicolumn{1}{c}{\centering {Block\_2}} & \includegraphics[width=3.06cm,height=2.3cm]{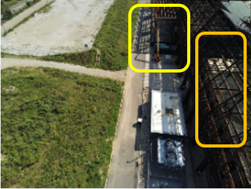} & \includegraphics[width=3.06cm,height=2.3cm]{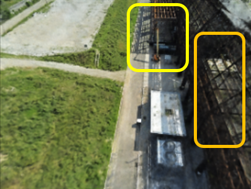} & \includegraphics[width=3.06cm,height=2.3cm]{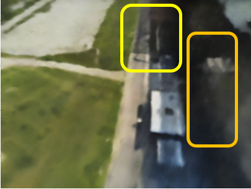} & \includegraphics[width=3.06cm,height=2.3cm]{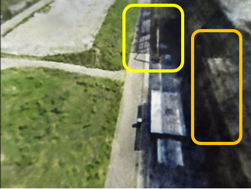}\\
        
        \multicolumn{1}{c}{\centering {Block\_3}} & \includegraphics[width=3.06cm,height=2.3cm]{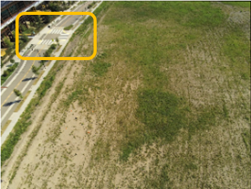} & \includegraphics[width=3.06cm,height=2.3cm]{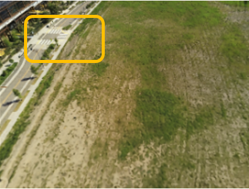} & \includegraphics[width=3.06cm,height=2.3cm]{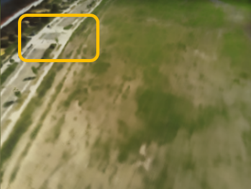} & \includegraphics[width=3.06cm,height=2.3cm]{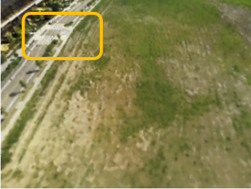} \\
        
        \multicolumn{1}{c}{\centering {Block\_4}} & \includegraphics[width=3.06cm,height=2.3cm]{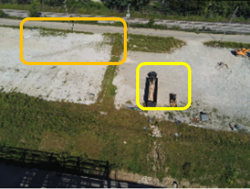} & \includegraphics[width=3.06cm,height=2.3cm]{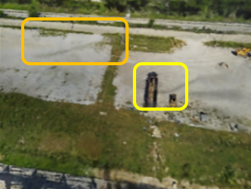} & \includegraphics[width=3.06cm,height=2.3cm]{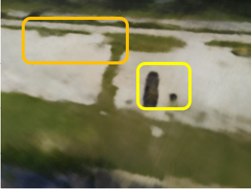} & \includegraphics[width=3.06cm,height=2.3cm]{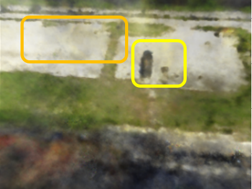} \\
        \hline
    \end{tabular}
    }
\end{table*}

\subsection{Scene Reconstruction in Building Scene}
We undertook a comprehensive analysis of the partitioned sub-block utilizing varying methods over the course of 5,000 and 30,000 iterations, recording the evaluative metrics for each specific sub-block, as detailed in Table \ref{tab:building_eval_table}. The statistical outcomes stemming from the 5,000 iteration mark for each sub-block within the Building Scene delineate a discernible trend: the average values for each metric within each sub-scenario of our model consistently outperform those of competing models, particularly in the context of a relatively limited iteration count. Additionally, the results derived from the 30,000 iteration milestone for each sub-block reveal a continuation of this trend. While our model's average metric scores continue to outshine those of other models across various sub-scenes, the gap between our model's evaluation indices and those of alternative models begins to narrow.
Upon close examination of the evaluation indicators pertaining to the 5,000 iteration comparison for each sub-block, a nuanced pattern emerges. Amongst these indicators, only the Structural Similarity Index (SSIM) of Block\_2 in our model falls below that of Instant-NGP. In contrast, the evaluation metrics of the remaining sub-scenes within our model surpass those of its counterparts. Analogously, when checking the evaluation indicators following 30,000 iterations, a parallel pattern is discerned. Here, solely the Peak Signal-to-Noise Ratio (PSNR) of Block$\_$2 within our model registers lower than the corresponding value in MipNeRF. The remaining sub-scene evaluation indices within our model, however, consistently demonstrate superiority over other models.

Through comparative analyses presented in Tables \ref{tab:building_images_method_comparison_5k} and \ref{tab:building_images_method_comparison_30k}, a distinct pattern emerges: our model consistently demonstrates superior performance across each sub-scene at the 5,000 iteration juncture. Notably, when juxtaposed against alternative models, our framework excels in capturing finer intricacies, even within a truncated number of iterations. Furthermore, within each sub-block evaluated at the 30,000 iteration milestone, our model exhibits pronounced proficiency in rendering shadow details with heightened fidelity, effectively reconstructing diminutive elements with impressive accuracy.

\begin{table*} %table environment, replace [] with h! The effect is the same
\centering % Indicates centered
\captionsetup{justification=raggedright,singlelinecheck=false}
\renewcommand{\arraystretch}{1.2}
\caption{Comparative Performance Evaluation of Various Methods in the Rubble Scene. Each model undergoes iterations of 5,000 and 30,000 within individual sub-scenes.} 
    \label{tab:Rubble_eval_table}
    \begin{tabular}{c c c c c c c c}
    \hline
       \multirow{2}{*}{\textbf{Methods}}  & \textbf{Number of}  & \multirow{2}{*}{\textbf{Metric}} & \textbf{Block\_1} & \textbf{Block\_2} & \textbf{Block\_3} & \textbf{Block\_4} & \textbf{Average} 
       \\
       ~ & \textbf{iterations} & ~ & \textbf{491 imgs} & \textbf{339 imgs} & \textbf{359 imgs} & \textbf{489 imgs} & \textbf{tot. 1678 imgs} 
       \\
    \hline
& & PSNR \scriptsize{$\uparrow$} & 18.388 & 18.433 & 21.146 & 18.730 & \textbf{19.174}  \\
Drone-NeRF (Ours) & 5k & SSIM \scriptsize{$\uparrow$} & 0.474 & 0.491 & 0.511 & 0.447 & \textbf{0.481}  \\
& & LPIPS \scriptsize{$\downarrow$} & 0.514 & 0.497 & 0.532 & 0.584 & \textbf{0.532}   \\
\hline       
& & PSNR \scriptsize{$\uparrow$} & 18.297 & 18.300 & 19.541 & 17.796 & 18.483  \\
MipNeRF & 5k & SSIM \scriptsize{$\uparrow$} & 0.417 & 0.403 & 0.396 & 0.373 & 0.397  \\
& & LPIPS \scriptsize{$\downarrow$} & 0.695 & 0.723 & 0.715 & 0.741 & 0.718  \\
\hline       
& & PSNR \scriptsize{$\uparrow$} & 17.506 & 17.745 & 17.527 & 17.973 & 17.687 \\
Instant-NGP & 5k & SSIM \scriptsize{$\uparrow$} & 0.302 & 0.332 & 0.385 & 0.377 & 0.349  \\
& & LPIPS \scriptsize{$\downarrow$} & 0.553 & 0.528 & 0.661 & 0.641 & 0.595   \\
\hline
\hline
& & PSNR \scriptsize{$\uparrow$} & 18.820 & 19.278 & 21.210 & 18.740 & 19.512  \\
Drone-NeRF (Ours) & 30k & SSIM \scriptsize{$\uparrow$} & 0.501 & 0.547 & 0.570 & 0.494 & \textbf{0.528}  \\
& & LPIPS \scriptsize{$\downarrow$} & 0.510 & 0.473 & 0.446 & 0.527 & \textbf{0.489}   \\
\hline       
& & PSNR \scriptsize{$\uparrow$} & 19.566 & 19.338 & 21.418 & 19.572 & \textbf{19.973}  \\
MipNeRF & 30k & SSIM \scriptsize{$\uparrow$} & 0.444 & 0.443 & 0.440 & 0.405 & 0.433  \\
& & LPIPS \scriptsize{$\downarrow$} & 0.646 & 0.650 & 0.652 & 0.647 & 0.649   \\
\hline       
& & PSNR \scriptsize{$\uparrow$} & 18.834 & 17.731 & 17.100 & 18.626 & 18.073  \\
Instant-NGP & 30k & SSIM \scriptsize{$\uparrow$} & 0.442 & 0.333 & 0.382 & 0.412 & 0.392  \\
& & LPIPS \scriptsize{$\downarrow$} & 0.653 & 0.588 & 0.751 & 0.635 & 0.657   \\
    \hline
   \end{tabular}
\end{table*}

\begin{table*}
\captionsetup{justification=justified,singlelinecheck=false}
    \caption{Comparative Analysis of Methods within the Rubble Sub-Scene over 5,000 Iterations. The first column lists the names of the segmented scenes. The second column represents the Ground Truth reference. The subsequent three columns offer a comparison of outcomes from the three models. Highlighted orange and yellow boxes indicate the specific targets selected for comparison within the Rubble Scene.}
    \label{tab:Rubble_images_method_comparison_5k}
    \centering
    \renewcommand{\arraystretch}{1.5}
    % \small % Set text size to \small
    % \scriptsize
    {\footnotesize 
    \begin{tabular}{m{1cm}<{\centering}m{3.1cm}<{\centering}m{3.1cm}<{\centering}m{3.1cm}<{\centering}m{3.1cm}<{\centering}m{3.1cm}}
        \hline
        \textbf{Scene} & \textbf{Rubble} & \multicolumn{3}{c}{\textbf{Methods}} \\
        \cline{3-5}
        \textbf{Block2*2} & \textbf{Ground Truth} & \textbf{Drone-NeRF (Ours)} & \textbf{MipNeRF} & \textbf{Instant-NGP} \\
        \hline
        \\[-3.0ex]
        \multicolumn{1}{c}{\centering {Block\_1}} &  \includegraphics[width=3.06cm,height=2.3cm]{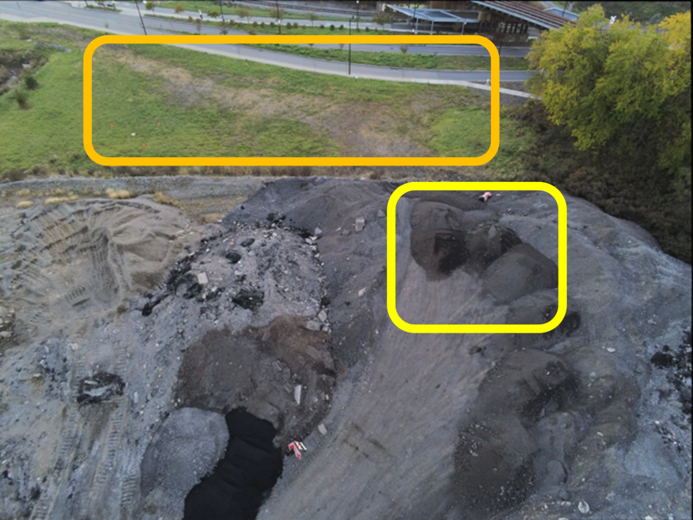}  &  \includegraphics[width=3.06cm,height=2.3cm]{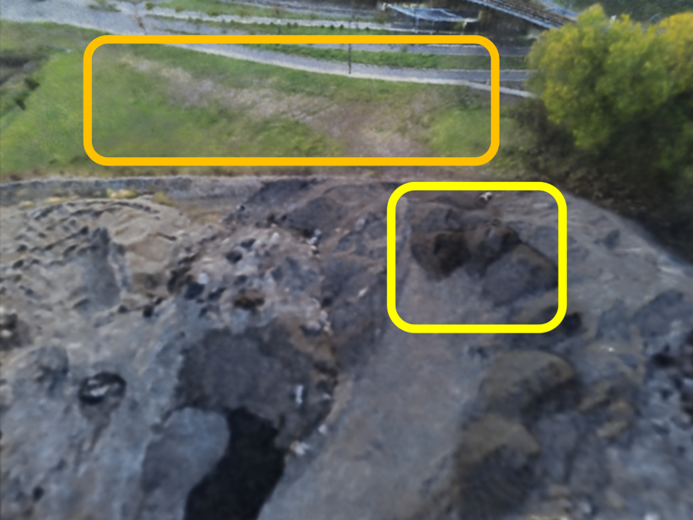}  & \includegraphics[width=3.06cm,height=2.3cm]{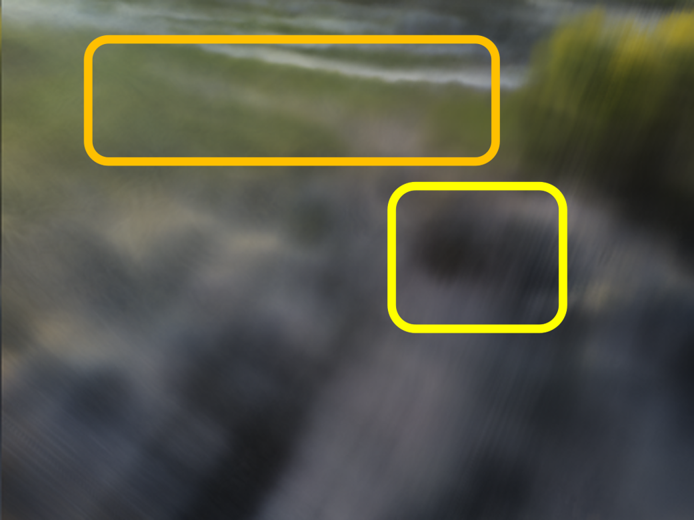}  & \includegraphics[width=3.06cm,height=2.3cm]{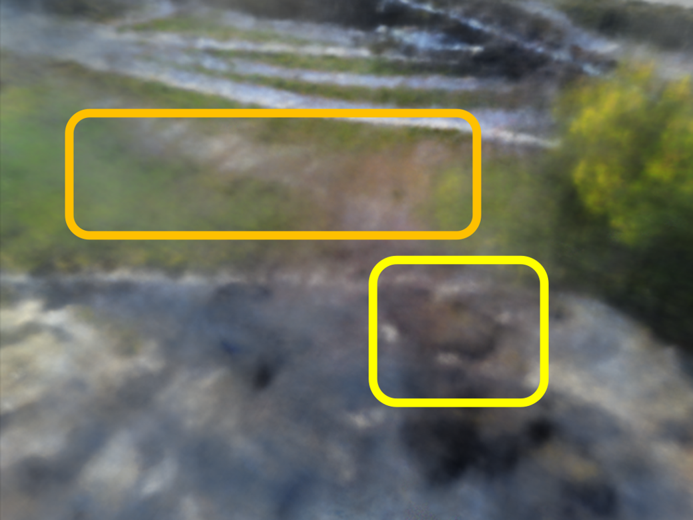} \\
        
        \multicolumn{1}{c}{\centering {Block\_2}} & \includegraphics[width=3.06cm,height=2.3cm]{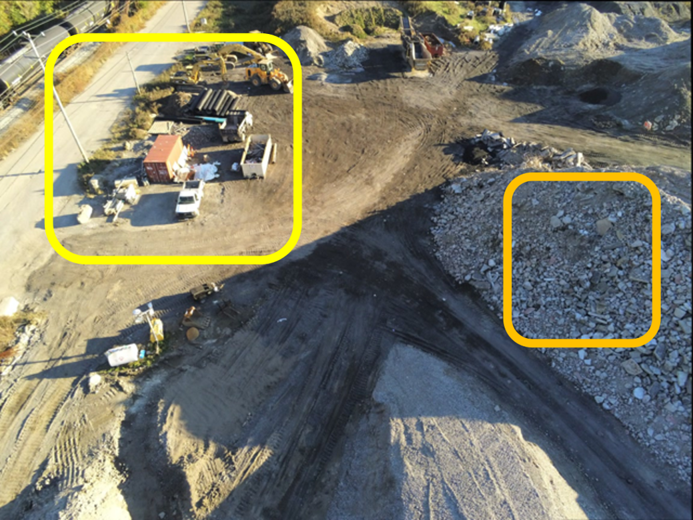} & \includegraphics[width=3.06cm,height=2.3cm]{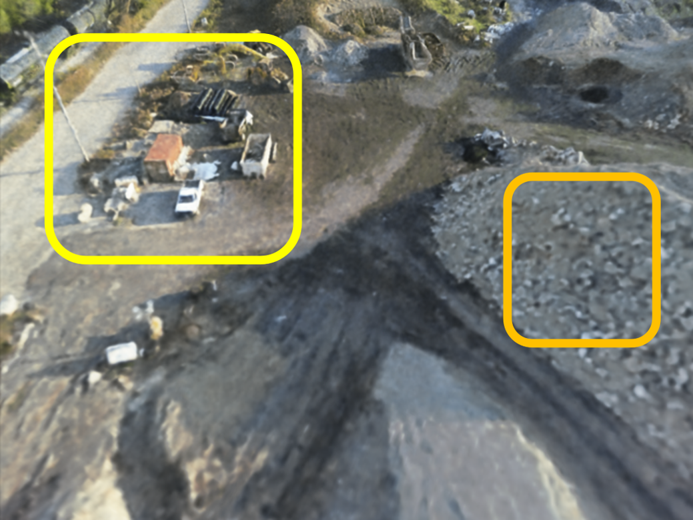} & \includegraphics[width=3.06cm,height=2.3cm]{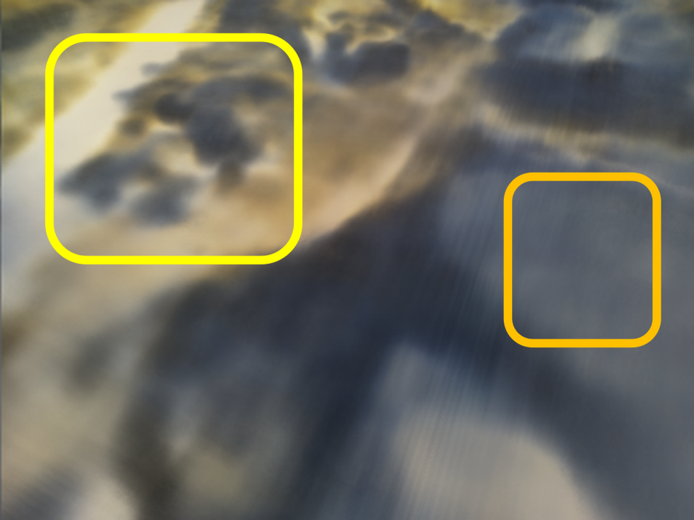} & \includegraphics[width=3.06cm,height=2.3cm]{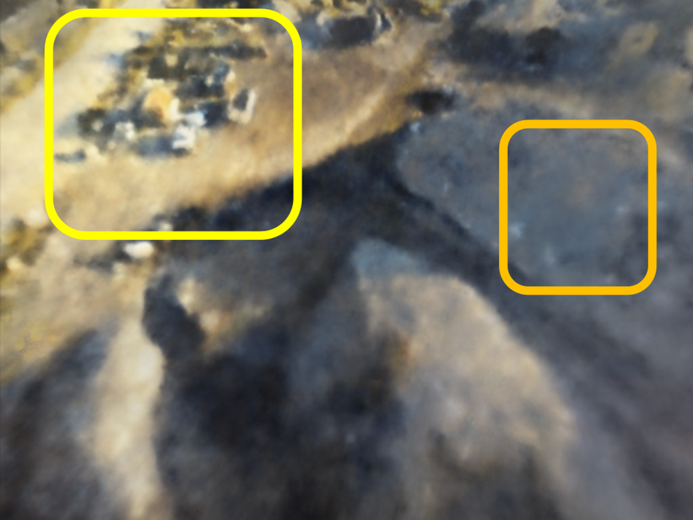}\\
        
        \multicolumn{1}{c}{\centering {Block\_3}} & \includegraphics[width=3.06cm,height=2.3cm]{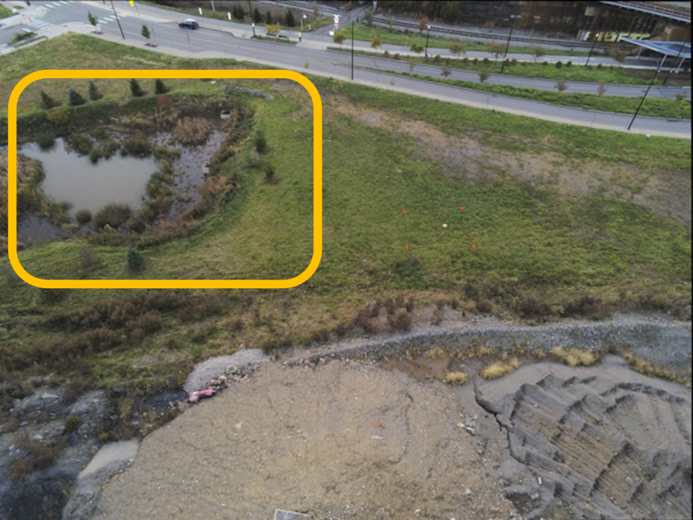} & \includegraphics[width=3.06cm,height=2.3cm]{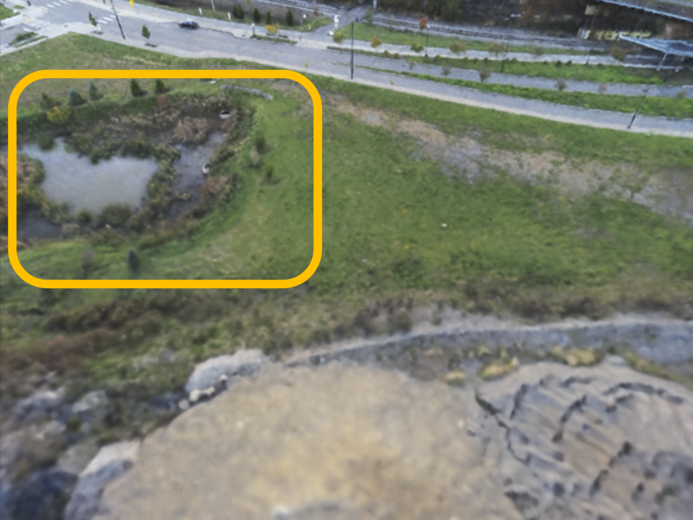} & \includegraphics[width=3.06cm,height=2.3cm]{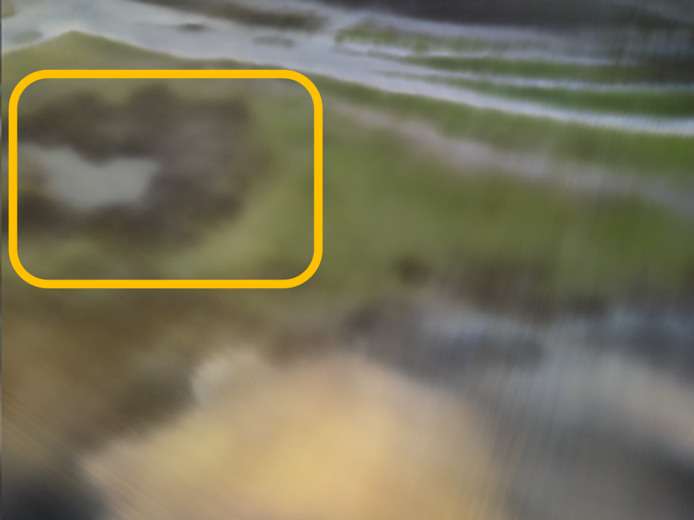} & \includegraphics[width=3.06cm,height=2.3cm]{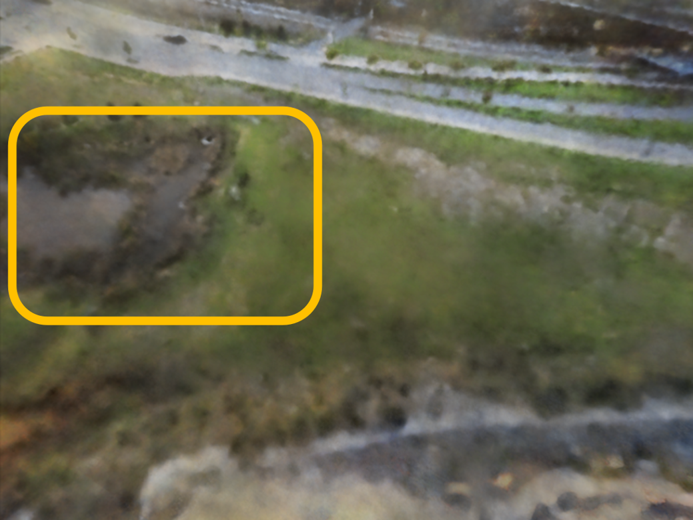} \\
        
        \multicolumn{1}{c}{\centering {Block\_4}} & \includegraphics[width=3.06cm,height=2.3cm]{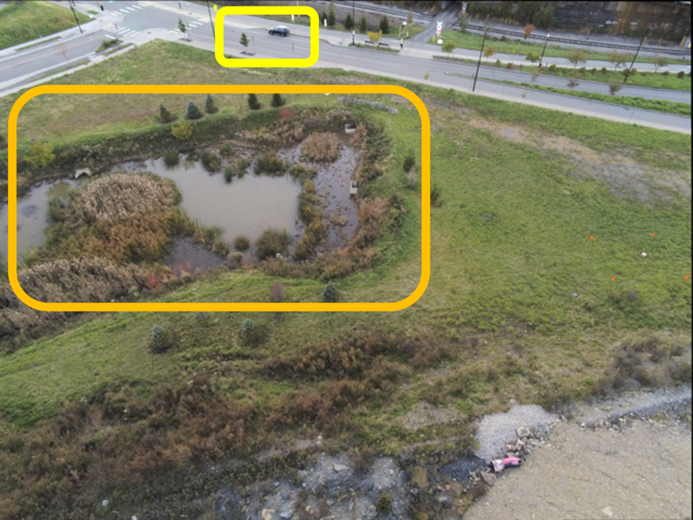} & \includegraphics[width=3.06cm,height=2.3cm]{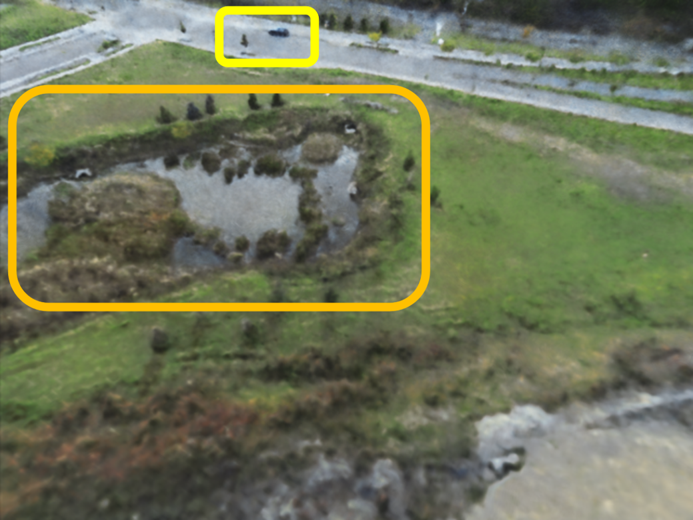} & \includegraphics[width=3.06cm,height=2.3cm]{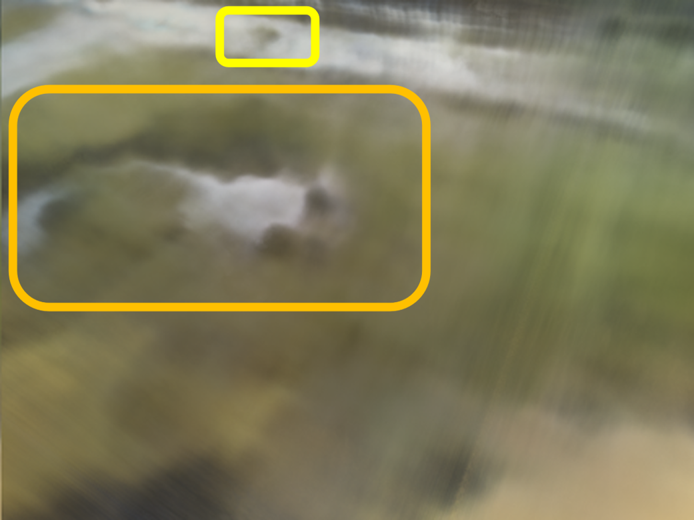} & \includegraphics[width=3.06cm,height=2.3cm]{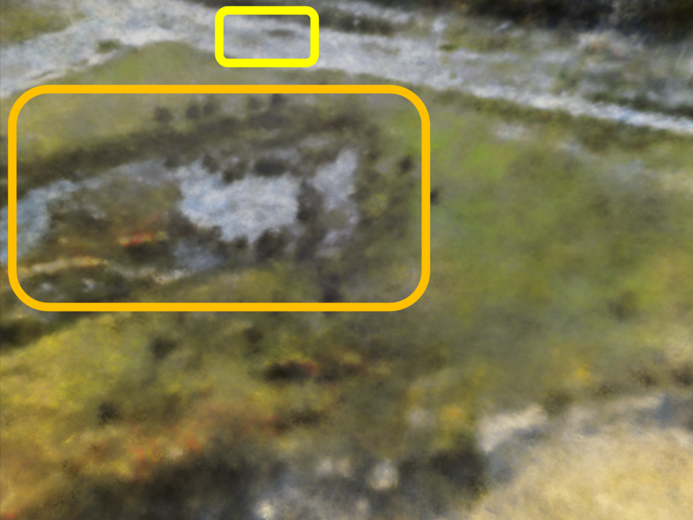} \\
        \hline
    \end{tabular}
    }
\end{table*}

\begin{table*}
\captionsetup{justification=justified,singlelinecheck=false}
    \caption{Comparative Analysis of Methods within the Rubble Sub-Scene over 30,000 Iterations. The first column lists the names of the segmented scenes. The second column represents the Ground Truth reference. The subsequent three columns offer a comparison of outcomes from the three models. Highlighted orange and yellow boxes indicate the specific targets selected for comparison within the Rubble Scene. }
    \label{tab:Rubble_images_method_comparison_30k}
    \centering
    \renewcommand{\arraystretch}{1.5}
    % \small % Set text size to \small
    % \scriptsize
    {\footnotesize 
    \begin{tabular}{m{1cm}<{\centering}m{3.1cm}<{\centering}m{3.1cm}<{\centering}m{3.1cm}<{\centering}m{3.1cm}<{\centering}m{3.1cm}}
        \hline
        \textbf{Scene} & \textbf{Rubble} & \multicolumn{3}{c}{\textbf{Methods}} \\
        \cline{3-5}
        \textbf{Block2*2} & \textbf{Ground Truth} & \textbf{Drone-NeRF (Ours)} & \textbf{MipNeRF} & \textbf{Instant-NGP} \\
        \hline
        \\[-3.0ex]
        \multicolumn{1}{c}{\centering {Block\_1}} &  \includegraphics[width=3.06cm,height=2.3cm]{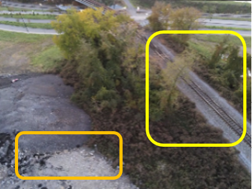}  &  \includegraphics[width=3.06cm,height=2.3cm]{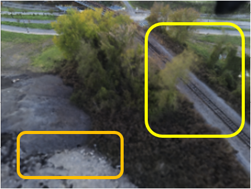}  & \includegraphics[width=3.06cm,height=2.3cm]{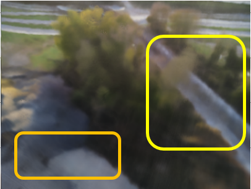}  & \includegraphics[width=3.06cm,height=2.3cm]{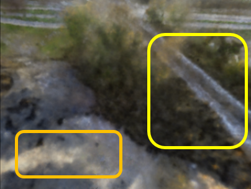} \\
        
        \multicolumn{1}{c}{\centering {Block\_2}} & \includegraphics[width=3.06cm,height=2.3cm]{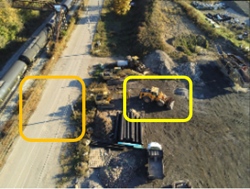} & \includegraphics[width=3.06cm,height=2.3cm]{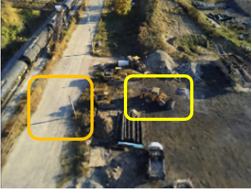} & \includegraphics[width=3.06cm,height=2.3cm]{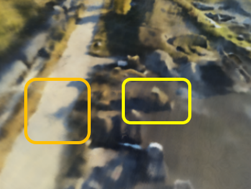} & \includegraphics[width=3.06cm,height=2.3cm]{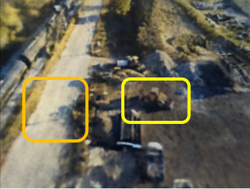}\\
        
        \multicolumn{1}{c}{\centering {Block\_3}} & \includegraphics[width=3.06cm,height=2.3cm]{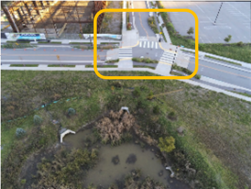} & \includegraphics[width=3.06cm,height=2.3cm]{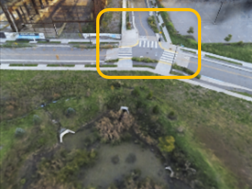} & \includegraphics[width=3.06cm,height=2.3cm]{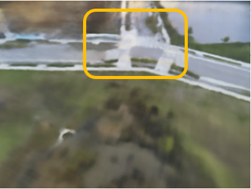} & \includegraphics[width=3.06cm,height=2.3cm]{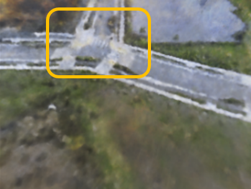} \\
        
        \multicolumn{1}{c}{\centering {Block\_4}} & \includegraphics[width=3.06cm,height=2.3cm]{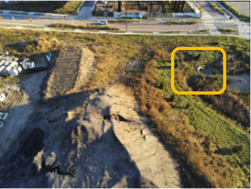} & \includegraphics[width=3.06cm,height=2.3cm]{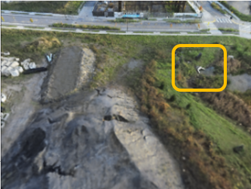} & \includegraphics[width=3.06cm,height=2.3cm]{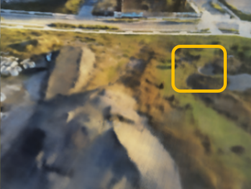} & \includegraphics[width=3.06cm,height=2.3cm]{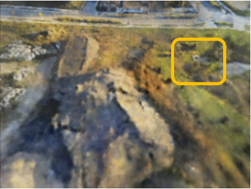} \\
        \hline
    \end{tabular}
    }
\end{table*}

Upon closer examination of the sub-scenes, we have several key observations. In Block$\_$1, assessed at the 5,000 iteration, our model adeptly accentuates the textural nuances of the crosswalk. Notably, Block$\_$2 showcases enhanced reconstruction of projecting architectural facets, accompanied by a more efficient convergence and reconstruction of vehicles within the parking area. Block$\_$3 evinces our model's superiority in delineating lawn details, a facet where alternative models fall short. In Block$\_$4, our model expeditiously reconstructs small arboreal elements, a task wherein the other models exhibit comparatively lesser efficacy.
Moving to the 30000 iteration assessments, distinctive patterns continue to surface. In Block$\_$1, a street light, observable in the ground truth, is adeptly reconstructed by our model, displaying an elevated level of fidelity. Additionally, our model excels in capturing the minutiae of billboards within Block$\_$1. Within the context of Block$\_$2, our model proficiently captures shadow-related nuances that closely resemble those present in real images. Notably, our model sustains its aptitude in expressing details within darker segments of the imagery. In the upper left corner of Block$\_$3, our model successfully reconstitutes the intricacies of the zebra crossing, affirming its precision in handling intricate features. The upper left corner of Block$\_$4 exhibits our model's proficiency in replicating the road imprint, demonstrating an alignment with the characteristics of the actual image. Noteworthy is our model's ability to discern and convey finer facets of the depicted truck, a domain wherein alternative models falter.

In summary, within the same 30,000 iteration training framework, our model showcases superior convergence speed and adeptness in capturing finer details within the Building Scene, including those residing within darker regions. This cumulative proficiency underscores our model's efficacy in both rapid convergence and comprehensive detail portrayal.

\subsection{Scene Reconstruction in Rubble Scene}
We continue our examination of diverse models across varying iterations for each sub-block within the Rubble Scene. The metrics for the comprehensive assessment of each sub-block are elucidated in Table \ref{tab:Rubble_eval_table}. Drawing insights from the statistical outcomes at the 5,000 iteration milestone, a conspicuous trend emerges: our model consistently outperforms its counterparts across every sub-block, manifesting higher average metric values within the ambit of a truncated iteration count. Moreover, upon delving into the statistical findings derived from 30,000 iterations within the Rubble sub-block context, a noteworthy pattern endures. The average metric values for each sub-scene within our model maintain commendable performance even as the iteration count extends.
However, as the iteration count escalates, a marginal shift is discernible. Specifically, MipNeRF marginally surpasses our model in the Peak Signal-to-Noise Ratio (PSNR) metric, while across the spectrum of other metrics, our model consistently surpasses alternative models. When checking the evaluation indicators across each sub-block at the 5,000 iteration threshold, the merits of our model stand evident, outshining those of other models. Similarly, upon dissecting the evaluation metrics for each sub-block at the 30,000 iteration mark, a salient trend crystallizes: our model boasts superior evaluation scores across each sub-scenario when compared to its counterparts, barring a marginal decline in the PSNR metric in relation to MipNeRF.

\begin{table*}
    \captionsetup{justification=justified, singlelinecheck=false}
    \caption{Comparative Analysis of Position Encoding with Varied Highest Frequencies over 30,000 Iterations. The investigation involves assessing the impact of different frequencies on the scene, particularly focusing on intricate details of small targets. This process aids in selecting an optimal frequency for the scene. The highlighted orange box denotes the specific target object selected for comparison within the scene.}
    \label{tab:max_frequency_pos_encoding}
    \centering
    \renewcommand{\arraystretch}{1.5}
    % \small % Set text size to \small
    % \scriptsize
    {\footnotesize 
    \begin{tabular}{m{3.2cm}<{\centering}m{3.2cm}<{\centering}m{3.2cm}<{\centering}m{3.2cm}<{\centering}m{3.2cm}<{\centering}m{3.2cm}}
        \hline
        \textbf{Scene} & \multicolumn{4}{c}{\textbf{Rubble Block\_1}} \\
        \cline{2-5}
        \textbf{Ground Truth} & \textbf{Frequency 16} & \textbf{Frequency 64} & \textbf{Frequency 128} & \textbf{Frequency 256} \\
        \hline
        \\[-3.0ex]
        \includegraphics[width=3.2cm,height=1.72cm]{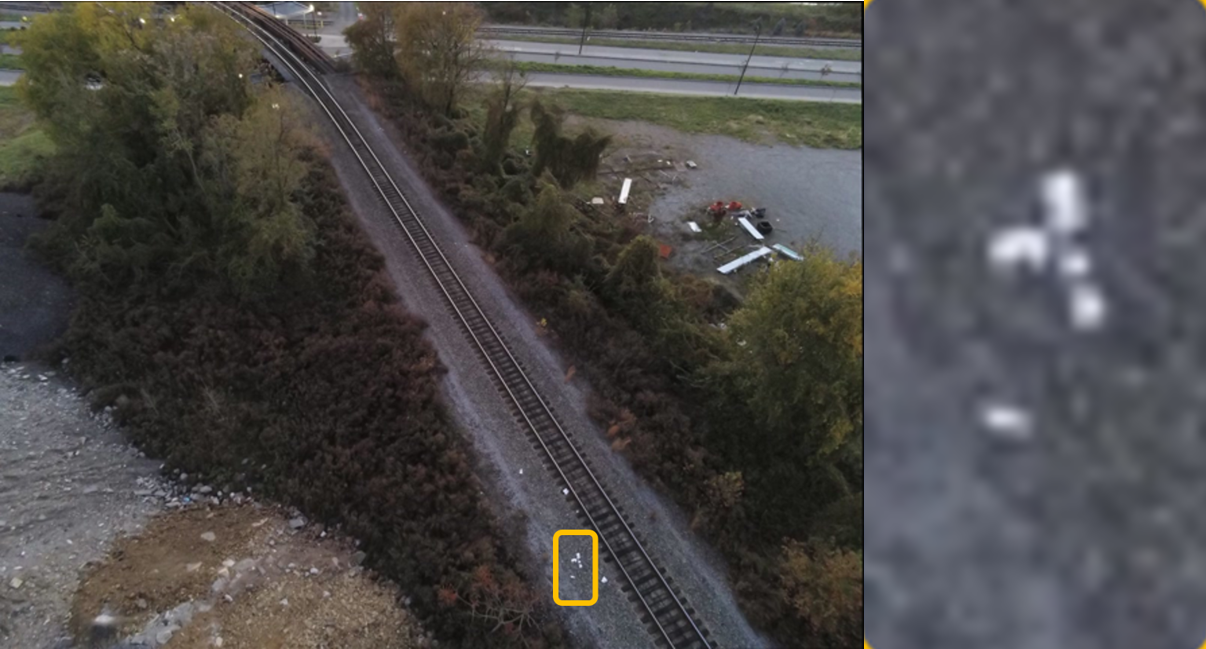} & \includegraphics[width=3.2cm,height=1.72cm]{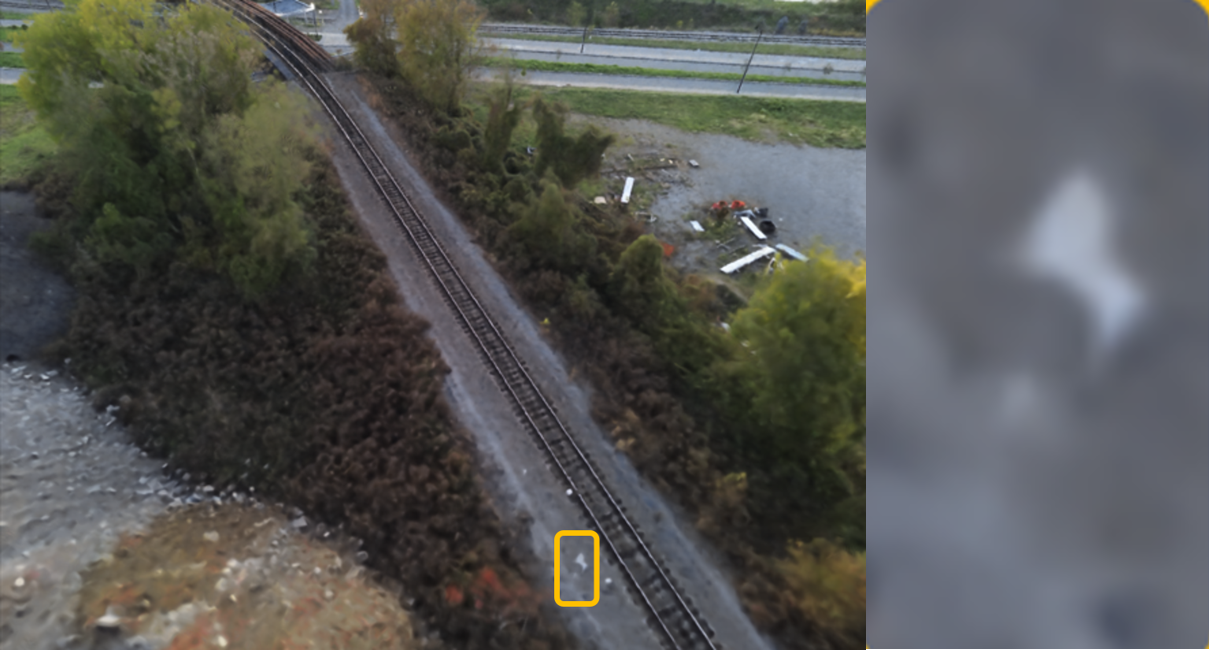} & \includegraphics[width=3.2cm,height=1.72cm]{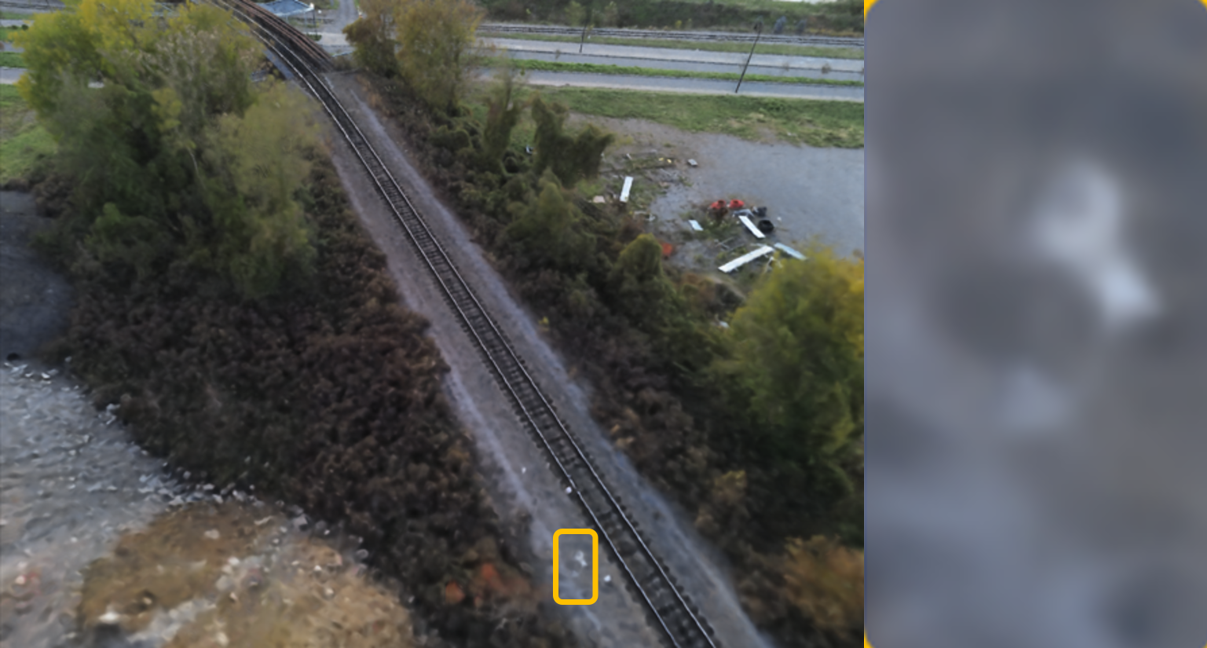} & \includegraphics[width=3.2cm,height=1.72cm]{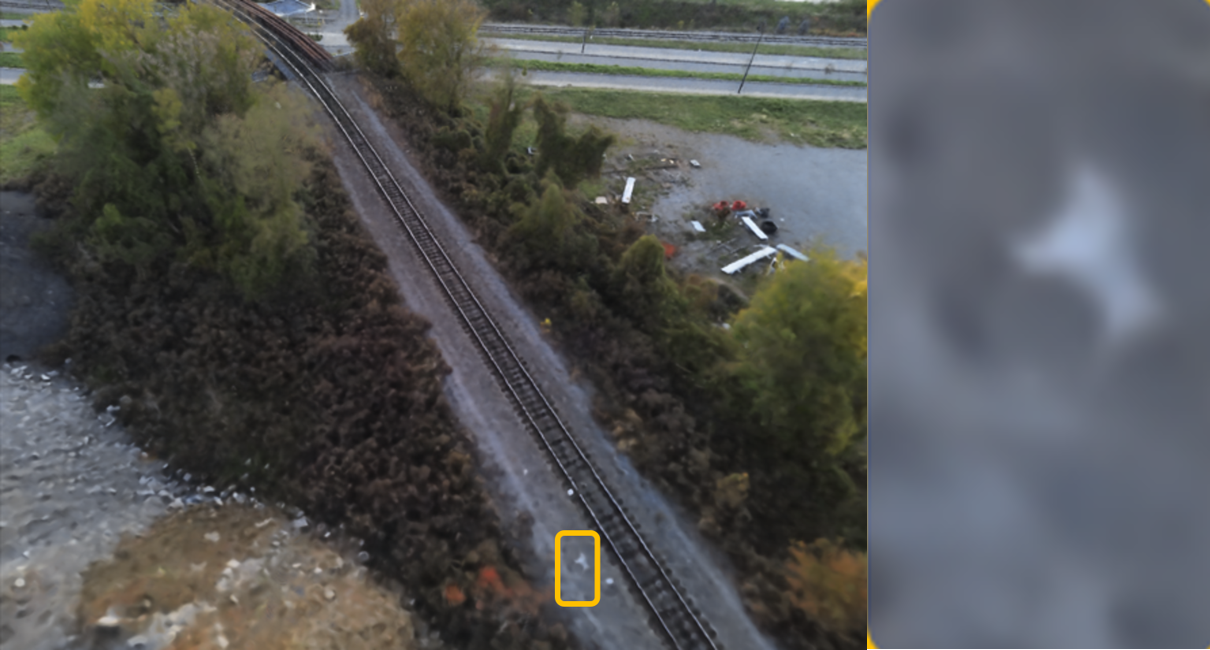} & \includegraphics[width=3.2cm,height=1.72cm]{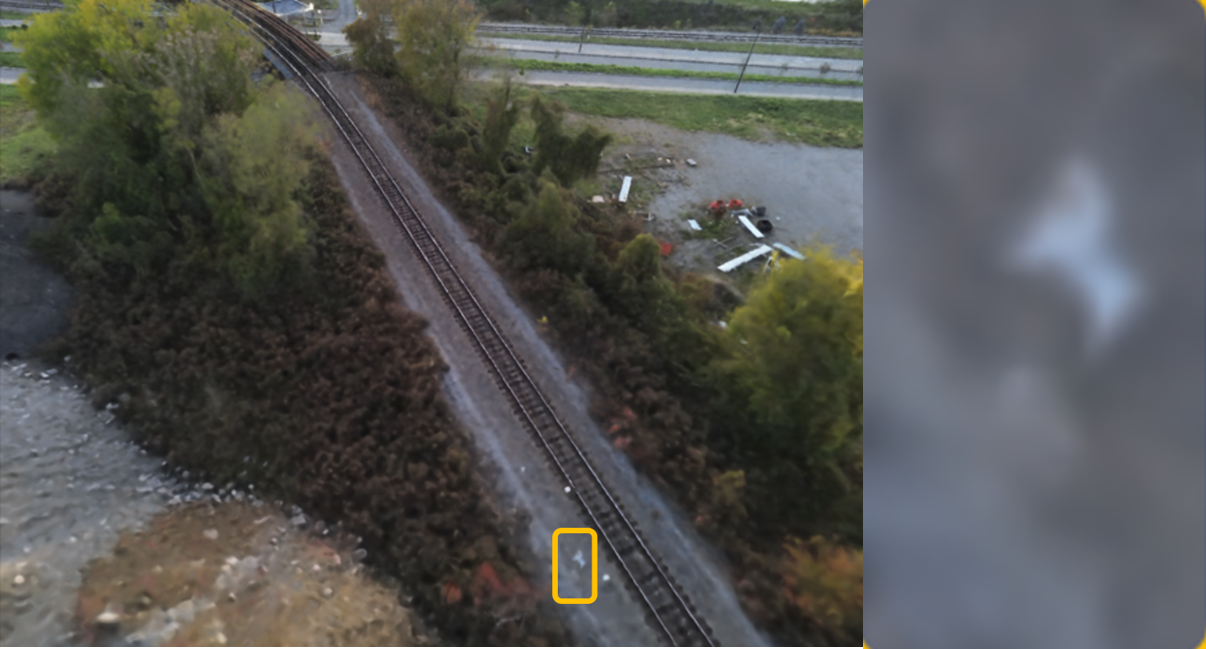} \\

        \multicolumn{1}{c}{\centering {PSNR \scriptsize{$\uparrow$}}} &  18.764  &  \textbf{18.819}  & 18.764  & 18.661 \\
        
        \multicolumn{1}{c}{\centering {SSIM \scriptsize{$\uparrow$}}} & 0.459 & \textbf{0.467} & 0.465 & 0.455\\
        
        \multicolumn{1}{c}{\centering {LPIPS \scriptsize{$\downarrow$}}} & 0.434 & \textbf{0.427} & 0.431 & 0.433 \\

        % \multicolumn{1}{c}{\centering {Ground Truth}} & frequency 16 & frequency 64 & frequency 128 & frequency 256 \\
        \hline
    \end{tabular}
    }
\end{table*}

Upon a thorough examination of the comparative analysis involving actual images, as shown in Tables \ref{tab:Rubble_images_method_comparison_5k} and \ref{tab:Rubble_images_method_comparison_30k}, discernible trends come to light. At the 5,000 iteration milestone, our model consistently produces superior outcomes within each sub-scene, accentuating its capacity to excel even within a truncated iteration count. Moreover, in contrast to alternative models, our framework showcases a remarkable ability to capture finer nuances in a condensed timeframe.
Furthermore, within each sub-block subjected to 30,000 iterations, our model continues to showcase proficiency in delineating shadow details and reconstructing intricate targets with remarkable accuracy. Notably, smaller targets also receive meticulous attention in our model's reconstruction.
The 5,000 iteration evaluations of each sub-block offer notable insights. Our model excels in enhancing path details on the lawn, demonstrating improved results even within a shorter iteration count. Notably, the effective reconstruction of the gravel pile within Block$\_$1 underscores our model's prowess. In Block$\_$2, rapid convergence in our model facilitates precise reconstruction of small houses and vehicles, enhancing contours. Moreover, our model distinguishes itself by delivering superior performance in handling smaller objects. Notably, within Block$\_$3 and Block$\_$4, our model excels in water area reconstruction, showcasing superiority over comparatively blurred alternatives. Moreover, in Block$\_$4, individual smaller trees and vehicles are distinctively expressed by our model.
Turning to the 30,000 iteration analyses of each sub-block, distinctive trends manifest. In Block$\_$1, our model excels in capturing the granularity of gravel, while alternatives exhibit blurriness. Additionally, our model shines in reconstructing tree branches, leaves, and railroad track details. Within Block$\_$2, the prominence of shadow details is a notable strength, while our model effectively reconstructs the forklift. In Block$\_$3, the zebra crossing is reconstructed with exceptional precision by our model. In Block$\_$4, our model's superior reconstruction of the pipeline entrance stands in contrast to the limitations of other models.
In sum, our model exhibits notable efficacy in swiftly converging to reconstruct scene details. Moreover, its prowess extends to capturing details of smaller objects.

\subsection{Efficiency Evaluation and Parameters Analysis}
We measured the time taken by our model to parallelize the processing of both the Building and Rubble scenarios subsequent to their division into distinct sub-blocks. Furthermore, we quantified the training time necessitated for iterations spanning 5,000 and 30,000 cycles respectively. This temporal evaluation extends to a comparative analysis involving two alternative models, MipNeRF and Instant-NGP, focusing on the training times required for these aforementioned scenarios. A comprehensive synthesis of these statistical outcomes is provided in Table \ref{tab:building_and_rubble_time_table}. Notably, the findings therein reveal a consistent trend: our model necessitates the shortest time duration, whether it entails 5,000 or 30,000 iterations, to traverse both the Building and Rubble Scenes.
This observation, when coupled with the insights gleaned from Tables \ref{tab:building_images_method_comparison_5k}, \ref{tab:building_images_method_comparison_30k}, \ref{tab:Rubble_images_method_comparison_5k}, and \ref{tab:Rubble_images_method_comparison_30k}, underscores a pivotal point. While ensuring expedient training, our model sustains commendable overall performance and an impressive attention to detail.

\begin{table} %table environment，[]Replace the middle with h! The effect is the same
\centering % Indicates centered
\captionsetup{justification=justified,singlelinecheck=false}
\renewcommand{\arraystretch}{1.2}
\caption{Comparative Time Consumption Analysis of Different Models with 5,000 and 30,000 Iterations in Building and Rubble Scenes.} 
    \label{tab:building_and_rubble_time_table}
    \footnotesize 
    \begin{tabular}{c c c c}
    \hline
       \multirow{2}{*}{\textbf{Scene}} & \multirow{2}{*}{\textbf{Methods}}  & \textbf{Average Time(min)} & \textbf{Average Time(min)} 
       \\
       ~ & ~ & \textbf{iterate 5k} & \textbf{iterate 30k} 
       \\
    \hline
\multirow{3}{*}{Building} & Drone-NeRF & \textbf{3.648} & \textbf{19.409} \\
~ & MipNeRF & 16.775 & 94.510 \\
~ & Instant-NGP  & 6.947 & 24.145 \\
\hline
\multirow{3}{*}{Rubble} & Drone-NeRF & \textbf{3.601} & \textbf{19.471} \\
~ & MipNeRF & 16.511 & 94.119 \\
~ & Instant-NGP  & 4.215 & 24.396 \\
    \hline
   \end{tabular}
\end{table}

\subsection{Ablation Study}
\subsubsection{The Comparison of Different Highest Frequency of Position Encoding}
The parameter for the maximum frequency of position encoding is systematically adjusted across values of 16, 64, 128, and 256. Through meticulous examination of its impact on high-frequency details, a discernible pattern emerges. Notably, when the uppermost frequency is set to 64, optimal performance is achieved in terms of evaluation metrics. This setting, when juxtaposed with the actual image, yields superior detail representation. It's imperative to acknowledge that as the maximum frequency increases, there is a concomitant rise in training time and GPU memory utilization, albeit within reasonable bounds. Guided by the need to balance high-frequency detail retention with efficient training and memory utilization, we choose the value of 64 as the pinnacle frequency for position encoding. This choice effectively accommodates the preservation of intricate high-frequency details while keeping training time and memory consumption within manageable limits. A comprehensive exposition of these findings is presented in Table \ref{tab:max_frequency_pos_encoding}.

\begin{table*} %table environment, replace [] with h! The effect is the same
\centering
\captionsetup{justification=justified,singlelinecheck=false}
\renewcommand{\arraystretch}{1.6}
\caption{Comparative Analysis of Varied Spatial Distortion Ranges for Block1 in the Rubble Scene. Each model undergoes 30,000 iterations within individual sub-scenes. Smaller spatial distortion ranges tend to induce more pronounced scene distortions. The highlighted orange box designates the specific target object under comparison within the scene.} 
    \label{tab:spatial_distortion_table}
    \begin{tabular}{m{0.8cm}<{\centering}m{1.2cm}<{\centering}m{0.8cm}<{\centering}m{0.8cm}<{\centering}m{0.8cm}<{\centering}m{0.8cm}<{\centering}m{3.1cm}<{\centering}m{4.8cm}<{\centering}m{0cm}}
    \hline
       \textbf{Scene} & \textbf{Contraction} & \textbf{Scale} & \textbf{PSNR}\scriptsize{$\uparrow$} & \textbf{SSIM}\scriptsize{$\uparrow$} & \textbf{LPIPS}\scriptsize{$\downarrow$} & \textbf{Spatial Distortion} & \textbf{Image ( left: Ground truth. right: Result. )}
       \\
    \hline
    \\[-3.0ex]
Rubble Block1 & L$_2$ norm & 0.5-1 & 13.555 & 0.229 & 0.647 & \includegraphics[width=1.89cm,height=1.8cm]{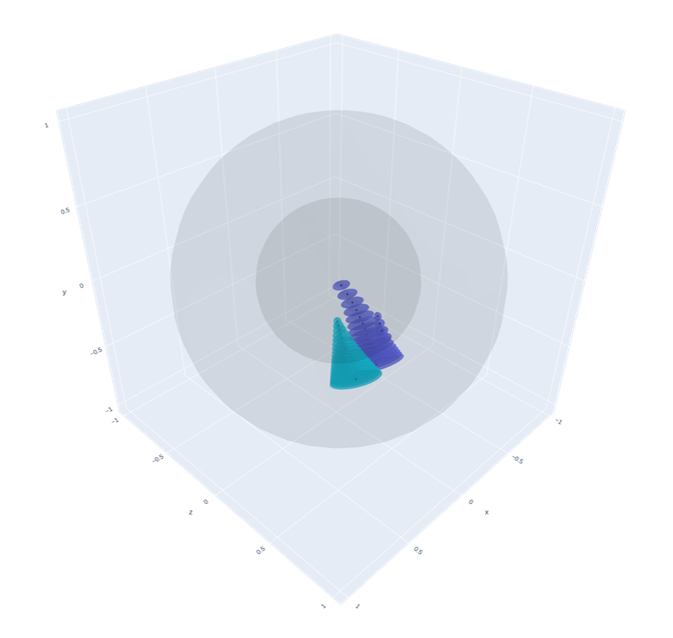} & \includegraphics[width=4.8cm,height=1.8cm]{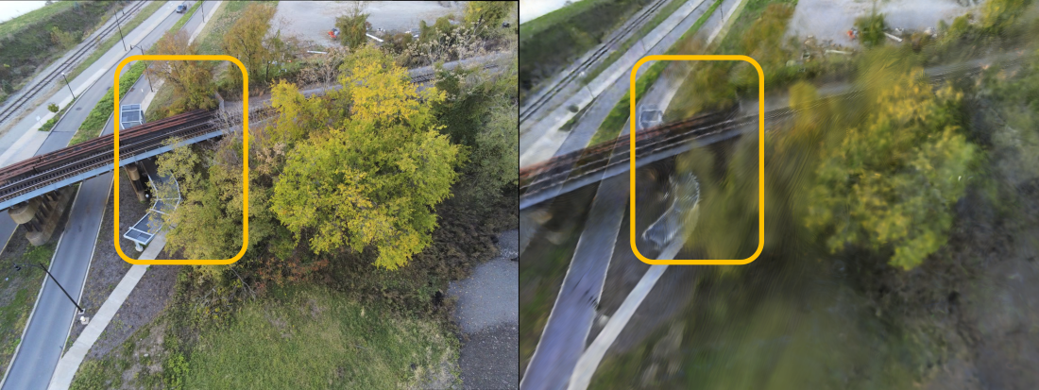} \\
Rubble Block1 & L$_\infty$ norm & 0.5-1 & 15.311 & 0.305 & 0.585 & \includegraphics[width=1.89cm,height=1.8cm]{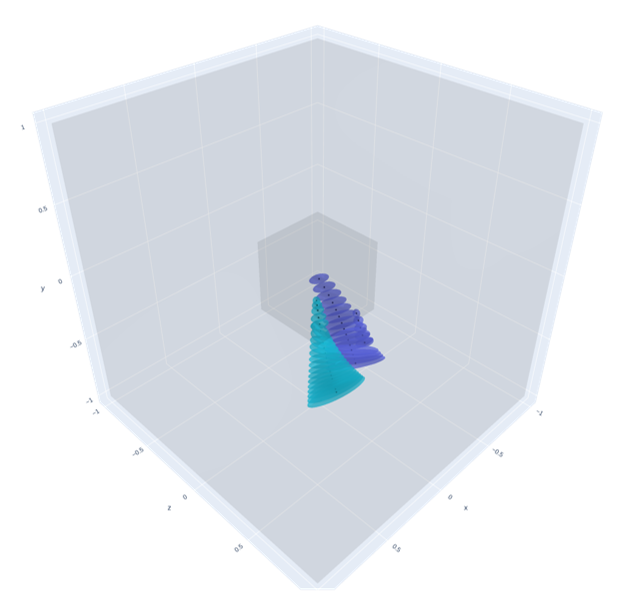} & \includegraphics[width=4.8cm,height=1.8cm]{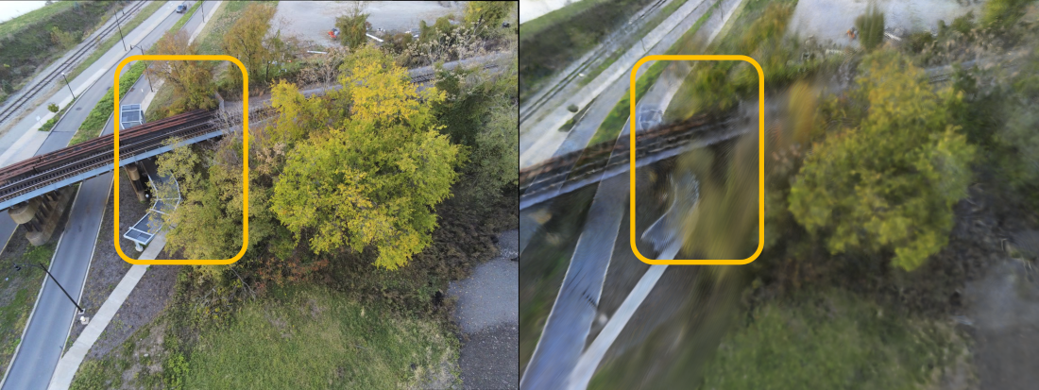} \\
Rubble Block1 & L$_2$ norm & 1.0-2.0 & 18.539 & 0.448 & 0.437 & \includegraphics[width=1.89cm,height=1.8cm]{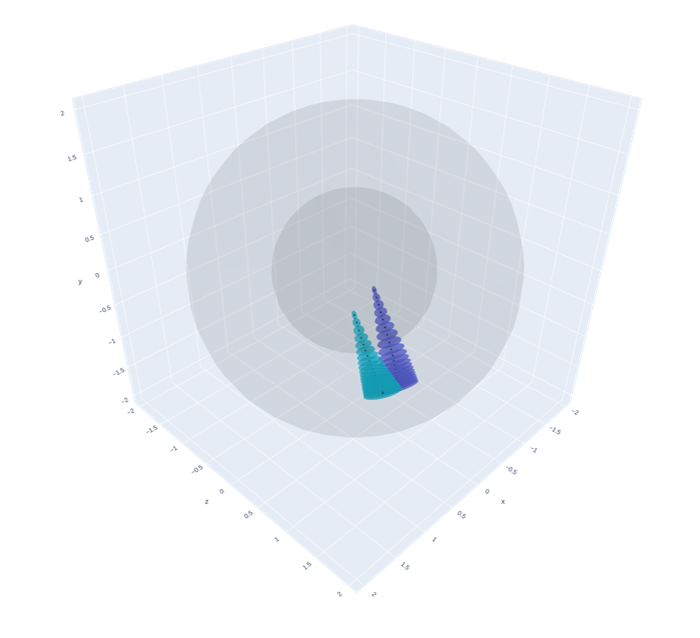} & \includegraphics[width=4.8cm,height=1.8cm]{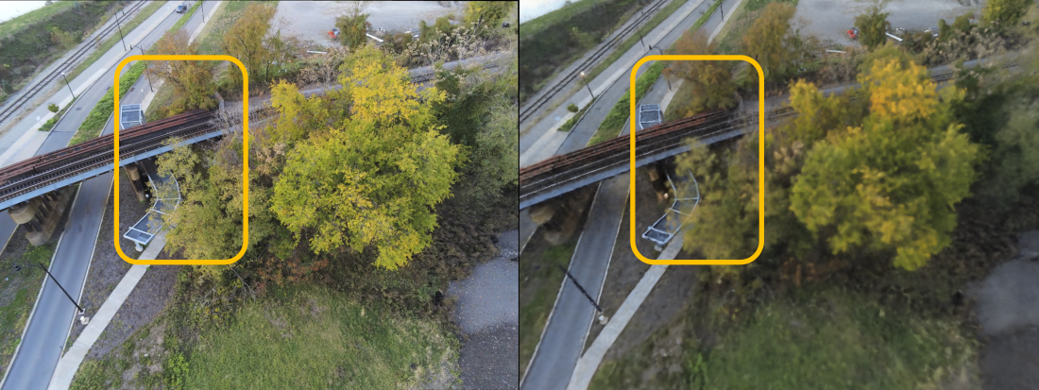} \\
Rubble Block1 & L$_\infty$ norm & 1.0-2.0 & \textbf{18.764} & 0.458 & \textbf{0.433} & \includegraphics[width=1.89cm,height=1.8cm]{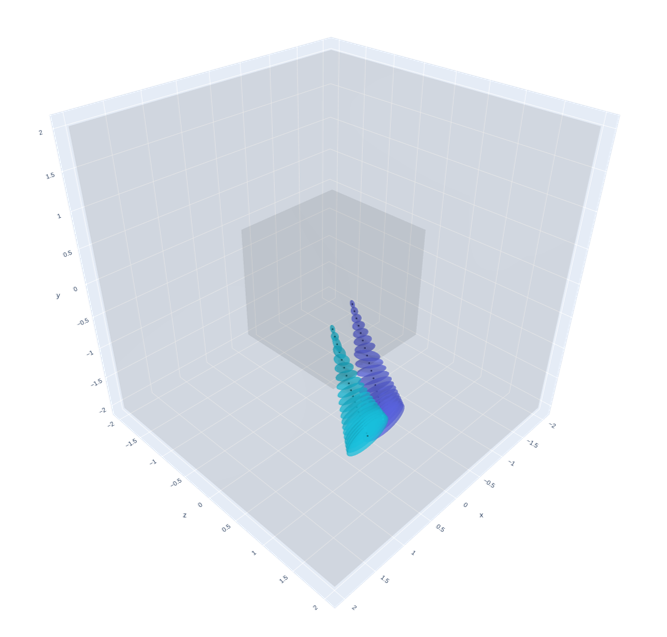} & \includegraphics[width=4.8cm,height=1.8cm]{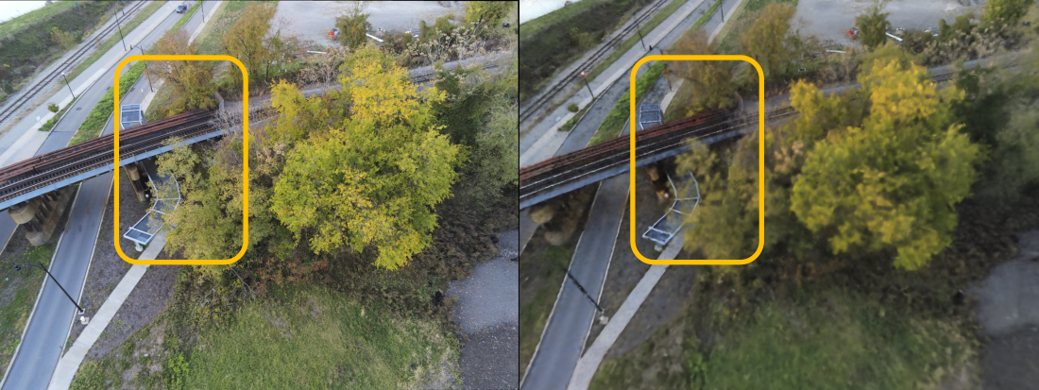} \\
Rubble Block1 & L$_2$ norm & 1.5-2.5 & 17.723 & 0.453 & 0.466 & \includegraphics[width=1.89cm,height=1.8cm]{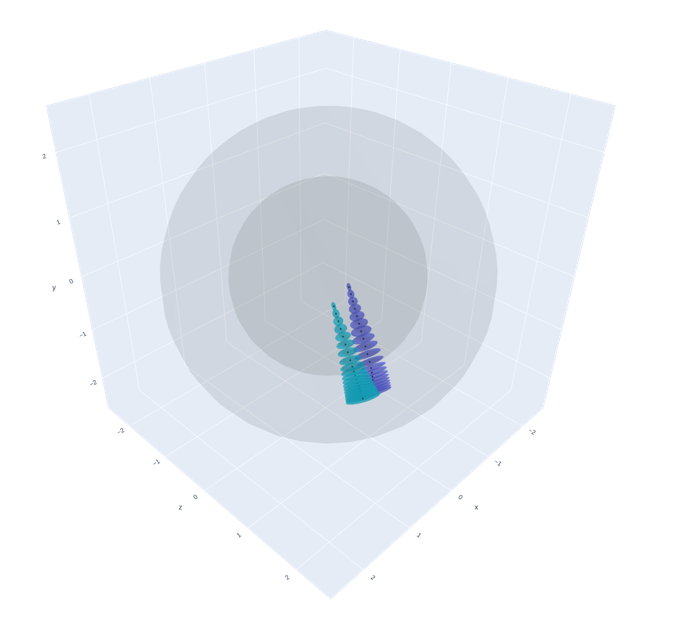} & \includegraphics[width=4.8cm,height=1.8cm]{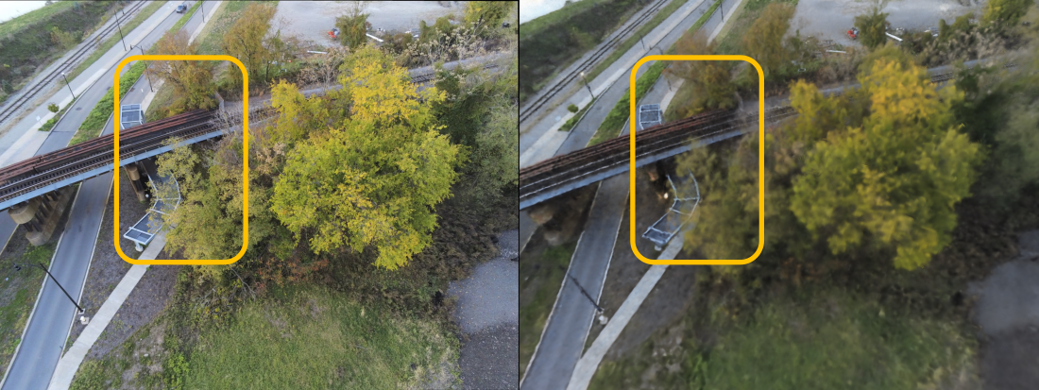} \\
Rubble Block1 & L$_\infty$ norm & 1.5-2.5 & 18.001 & \textbf{0.466} & 0.470 & \includegraphics[width=1.89cm,height=1.8cm]{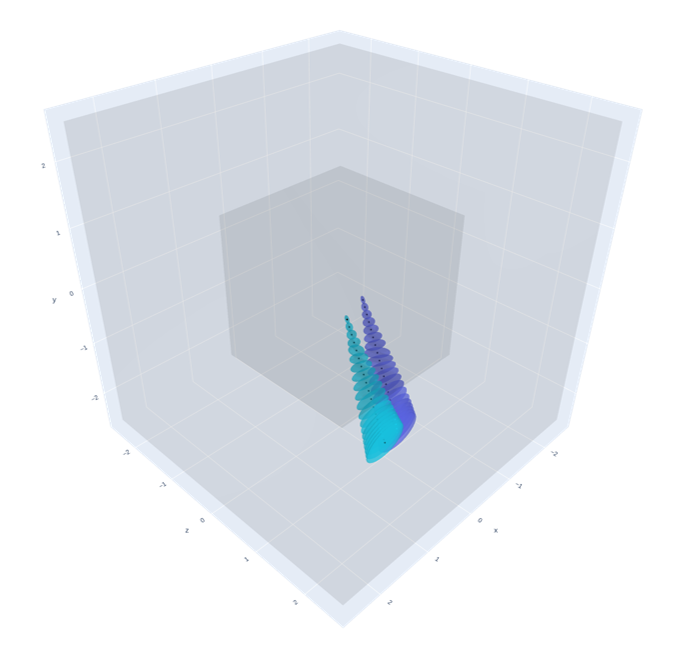} & \includegraphics[width=4.8cm,height=1.8cm]{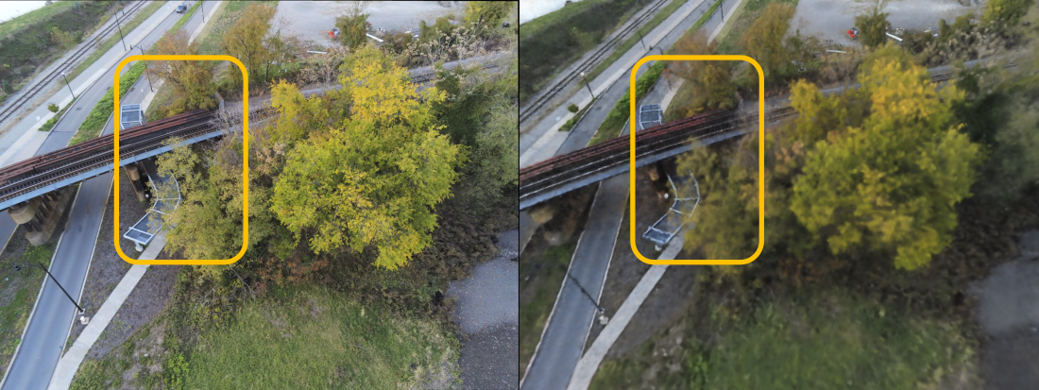} \\
    \hline
   \end{tabular}
\end{table*}

\subsubsection{The Comparison of Different Spatial Distortion Ranges}
We systematically modulate and subsequently check the scope of spatial distortions, thereby establishing distortion ranges across three tiers: 0.5 to 1, 1.0 to 2.0, and 1.5 to 2.5. Within this framework, scene rays are subjected to distortion and subsequently reprojected at regular intervals. Additionally, we juxtapose two distinct Contraction methods, specifically the L2 norm and the L$\infty$ norm. A comprehensive analysis of the outcomes reveals that varying spatial distortion ranges exhibit a limited influence on training time. The L$\infty$ norm aligns more seamlessly with the chunked post-subcube scene, underscoring its compatibility with the spatial attributes. Within the 0.5 to 1.0 range, conspicuous distortion and blurriness are evident across a substantial area. In comparison with the other two ranges, the 0.5 to 1.0 range distinctly exhibits more pronounced distortion, paralleled by diminished evaluation scores. Comparatively, the remaining two ranges better cater to the spatial characteristics of the scene, as the impact of spatial distortion is comparatively reduced, culminating in superior outcomes. The comparative analysis results are shown in Table \ref{tab:spatial_distortion_table}.
In the comparative analysis between the two ranges, namely 1.0 to 2.0 and 1.5 to 2.5, the former emerges as more favourable in terms of evaluation indices. As a result of this analysis, our model decisively opts for the 1.0 to 2.0 distortion range, a choice supported by its more beneficial alignment with the requirements of the spatial scene and superior performance.

\section{Conclusion}
In this work, we introduce a novel parallel modular approach designed to partition training sub-blocks for drone-view scenes while concurrently amalgamating the rendering outcomes of each individual sub-scene. This technique is tailored to facilitate the 3D reconstruction of expansive drone-captured scenes. Employing a NeRF-based framework that prioritizes rapid convergence, we efficiently accomplish scene training, yielding high-quality reconstructed scenes within a compressed timeframe. Notably, this approach deftly accounts for both local and global scene characteristics inherent to each sub-scene.
By implementing our method, we successfully surmount the array of challenges associated with large-scale drone-view scenarios. This not only attests to the method's efficiency but also assures the fidelity of the reconstruction process. Consequently, our method offers significant practical utility, particularly in domains such as 3D reconstruction based on drone imagery and digital surface models founded on depth maps.

 As the accuracy of obtained pose data plays a crucial role in influencing the quality of scene reconstruction, our future research includes integrating high-precision RTK modules in later stages to enhance the precision of pose estimation. Moreover, in the domain of rendering, the necessity to combine sub-scenes has constrained the speed of interactive applications. As part of our future research, we intend to draw insights from hierarchical display mechanisms. This approach involves consolidating and rendering only the sub-scenes within the current field of view. Sub-scenes beyond the immediate field of view will not be rendered, and those located at a distance will be presented at lower resolutions. This innovative strategy is anticipated to expedite display speed, ultimately leading to improved interactive experiences.

% reference:
\bibliographystyle{unsrt}
\bibliography{Mybib}

\begin{thebibliography}{10}

\bibitem{inzerillo2018image}
Laura Inzerillo, Gaetano Di~Mino, and Ronald Roberts.
\newblock Image-based 3d reconstruction using traditional and uav datasets for
  analysis of road pavement distress.
\newblock {\em Automation in Construction}, 96:457--469, 2018.

\bibitem{zhao2021structural}
Sizeng Zhao, Fei Kang, Junjie Li, and Chuanbo Ma.
\newblock Structural health monitoring and inspection of dams based on uav
  photogrammetry with image 3d reconstruction.
\newblock {\em Automation in Construction}, 130:103832, 2021.

\bibitem{shang2018real}
Zhexiong Shang and Zhigang Shen.
\newblock Real-time 3d reconstruction on construction site using visual slam
  and uav.
\newblock In {\em Construction Research Congress 2018}, pages 305--315, 2018.

\bibitem{jiang2020uav}
Weiguang Jiang, Ying Zhou, Lieyun Ding, Cheng Zhou, and Xiaodi Ning.
\newblock Uav-based 3d reconstruction for hoist site mapping and layout
  planning in petrochemical construction.
\newblock {\em Automation in Construction}, 113:103137, 2020.

\bibitem{snavely2006photo}
Noah Snavely, Steven~M Seitz, and Richard Szeliski.
\newblock Photo tourism: exploring photo collections in 3d.
\newblock In {\em ACM siggraph 2006 papers}, pages 835--846. 2006.

\bibitem{shen2013accurate}
Shuhan Shen.
\newblock Accurate multiple view 3d reconstruction using patch-based stereo for
  large-scale scenes.
\newblock {\em IEEE transactions on image processing}, 22(5):1901--1914, 2013.

\bibitem{xu20193d}
Haonan Xu, Junyi Hou, Lei Yu, and Shumin Fei.
\newblock 3d reconstruction system for collaborative scanning based on multiple
  rgb-d cameras.
\newblock {\em Pattern Recognition Letters}, 128:505--512, 2019.

\bibitem{vu2011high}
Hoang-Hiep Vu, Patrick Labatut, Jean-Philippe Pons, and Renaud Keriven.
\newblock High accuracy and visibility-consistent dense multiview stereo.
\newblock {\em IEEE transactions on pattern analysis and machine intelligence},
  34(5):889--901, 2011.

\bibitem{carr2001reconstruction}
Jonathan~C Carr, Richard~K Beatson, Jon~B Cherrie, Tim~J Mitchell, W~Richard
  Fright, Bruce~C McCallum, and Tim~R Evans.
\newblock Reconstruction and representation of 3d objects with radial basis
  functions.
\newblock In {\em Proceedings of the 28th annual conference on Computer
  graphics and interactive techniques}, pages 67--76, 2001.

\bibitem{mildenhall2021nerf}
Ben Mildenhall, Pratul~P Srinivasan, Matthew Tancik, Jonathan~T Barron, Ravi
  Ramamoorthi, and Ren Ng.
\newblock Nerf: Representing scenes as neural radiance fields for view
  synthesis.
\newblock {\em Communications of the ACM}, 65(1):99--106, 2021.

\bibitem{ost2021neural}
Julian Ost, Fahim Mannan, Nils Thuerey, Julian Knodt, and Felix Heide.
\newblock Neural scene graphs for dynamic scenes.
\newblock In {\em Proceedings of the IEEE/CVF Conference on Computer Vision and
  Pattern Recognition}, pages 2856--2865, 2021.

\bibitem{song2022implicit}
Boyang Song, Xiaoguang Hu, Jin Xiao, Guofeng Zhang, and Tianyou Chen.
\newblock Implicit neural refinement based multi-view stereo network with
  adaptive correlation.
\newblock {\em Image and Vision Computing}, 124:104511, 2022.

\bibitem{mari2022sat}
Roger Mar{\'\i}, Gabriele Facciolo, and Thibaud Ehret.
\newblock Sat-nerf: Learning multi-view satellite photogrammetry with transient
  objects and shadow modeling using rpc cameras.
\newblock In {\em Proceedings of the IEEE/CVF Conference on Computer Vision and
  Pattern Recognition}, pages 1311--1321, 2022.

\bibitem{noonan2021neuralplan}
John Noonan, Ehud Rivlin, and Hector Rotstein.
\newblock Neuralplan: Neural floorplan radiance fields for accelerated view
  synthesis.
\newblock {\em Image and Vision Computing}, 109:104148, 2021.

\bibitem{tancik2022block}
Matthew Tancik, Vincent Casser, Xinchen Yan, Sabeek Pradhan, Ben Mildenhall,
  Pratul~P Srinivasan, Jonathan~T Barron, and Henrik Kretzschmar.
\newblock Block-nerf: Scalable large scene neural view synthesis.
\newblock In {\em Proceedings of the IEEE/CVF Conference on Computer Vision and
  Pattern Recognition}, pages 8248--8258, 2022.

\bibitem{hao2022review}
YaNan Hao, YC~Tan, VC~Tai, XD~Zhang, EP~Wei, and SC~Ng.
\newblock Review of key technologies for warehouse 3d reconstruction.
\newblock {\em Journal of Mechanical Engineering and Sciences},
  16(3):9142--9156, 2022.

\bibitem{xu2022depth}
Yuhua Xu, Xiaoli Yang, Yushan Yu, Wei Jia, Zhaobi Chu, and Yulan Guo.
\newblock Depth estimation by combining binocular stereo and monocular
  structured-light.
\newblock In {\em Proceedings of the IEEE/CVF Conference on Computer Vision and
  Pattern Recognition}, pages 1746--1755, 2022.

\bibitem{horn1986variational}
Berthold~KP Horn and Michael~J Brooks.
\newblock The variational approach to shape from shading.
\newblock {\em Computer Vision, Graphics, and Image Processing},
  33(2):174--208, 1986.

\bibitem{woodham1979photometric}
Robert~J Woodham.
\newblock Photometric stereo: A reflectance map technique for determining
  surface orientation from image intensity.
\newblock In {\em Image understanding systems and industrial applications I},
  volume 155, pages 136--143. SPIE, 1979.

\bibitem{jin2005multi}
Hailin Jin, Stefano Soatto, and Anthony~J Yezzi.
\newblock Multi-view stereo reconstruction of dense shape and complex
  appearance.
\newblock {\em International Journal of Computer Vision}, 63:175--189, 2005.

\bibitem{seitz2006comparison}
Steven~M Seitz, Brian Curless, James Diebel, Daniel Scharstein, and Richard
  Szeliski.
\newblock A comparison and evaluation of multi-view stereo reconstruction
  algorithms.
\newblock In {\em 2006 IEEE computer society conference on computer vision and
  pattern recognition (CVPR'06)}, volume~1, pages 519--528. IEEE, 2006.

\bibitem{xi2022raymvsnet}
Junhua Xi, Yifei Shi, Yijie Wang, Yulan Guo, and Kai Xu.
\newblock Raymvsnet: Learning ray-based 1d implicit fields for accurate
  multi-view stereo.
\newblock In {\em Proceedings of the IEEE/CVF Conference on Computer Vision and
  Pattern Recognition}, pages 8595--8605, 2022.

\bibitem{rematas2022urban}
Konstantinos Rematas, Andrew Liu, Pratul~P Srinivasan, Jonathan~T Barron,
  Andrea Tagliasacchi, Thomas Funkhouser, and Vittorio Ferrari.
\newblock Urban radiance fields.
\newblock In {\em Proceedings of the IEEE/CVF Conference on Computer Vision and
  Pattern Recognition}, pages 12932--12942, 2022.

\bibitem{cai2022pix2nerf}
Shengqu Cai, Anton Obukhov, Dengxin Dai, and Luc Van~Gool.
\newblock Pix2nerf: Unsupervised conditional p-gan for single image to neural
  radiance fields translation.
\newblock In {\em Proceedings of the IEEE/CVF Conference on Computer Vision and
  Pattern Recognition}, pages 3981--3990, 2022.

\bibitem{athar2022rignerf}
ShahRukh Athar, Zexiang Xu, Kalyan Sunkavalli, Eli Shechtman, and Zhixin Shu.
\newblock Rignerf: Fully controllable neural 3d portraits.
\newblock In {\em Proceedings of the IEEE/CVF Conference on Computer Vision and
  Pattern Recognition}, pages 20364--20373, 2022.

\bibitem{krishnan2023novel}
Anoop Krishnan and Ajita Rattani.
\newblock A novel approach for bias mitigation of gender classification
  algorithms using consistency regularization.
\newblock {\em Image and Vision Computing}, page 104793, 2023.

\bibitem{xu2022surface}
Tianhan Xu, Yasuhiro Fujita, and Eiichi Matsumoto.
\newblock Surface-aligned neural radiance fields for controllable 3d human
  synthesis.
\newblock In {\em Proceedings of the IEEE/CVF Conference on Computer Vision and
  Pattern Recognition}, pages 15883--15892, 2022.

\bibitem{shao2022doublefield}
Ruizhi Shao, Hongwen Zhang, He~Zhang, Mingjia Chen, Yan-Pei Cao, Tao Yu, and
  Yebin Liu.
\newblock Doublefield: Bridging the neural surface and radiance fields for
  high-fidelity human reconstruction and rendering.
\newblock In {\em Proceedings of the IEEE/CVF Conference on Computer Vision and
  Pattern Recognition}, pages 15872--15882, 2022.

\bibitem{weng2022humannerf}
Chung-Yi Weng, Brian Curless, Pratul~P Srinivasan, Jonathan~T Barron, and Ira
  Kemelmacher-Shlizerman.
\newblock Humannerf: Free-viewpoint rendering of moving people from monocular
  video.
\newblock In {\em Proceedings of the IEEE/CVF Conference on Computer Vision and
  Pattern Recognition}, pages 16210--16220, 2022.

\bibitem{yang2022neumesh}
Bangbang Yang, Chong Bao, Junyi Zeng, Hujun Bao, Yinda Zhang, Zhaopeng Cui, and
  Guofeng Zhang.
\newblock Neumesh: Learning disentangled neural mesh-based implicit field for
  geometry and texture editing.
\newblock In {\em European Conference on Computer Vision}, pages 597--614.
  Springer, 2022.

\bibitem{yang2021learning}
Bangbang Yang, Yinda Zhang, Yinghao Xu, Yijin Li, Han Zhou, Hujun Bao, Guofeng
  Zhang, and Zhaopeng Cui.
\newblock Learning object-compositional neural radiance field for editable
  scene rendering.
\newblock In {\em Proceedings of the IEEE/CVF International Conference on
  Computer Vision}, pages 13779--13788, 2021.

\bibitem{liu2021editing}
Steven Liu, Xiuming Zhang, Zhoutong Zhang, Richard Zhang, Jun-Yan Zhu, and
  Bryan Russell.
\newblock Editing conditional radiance fields.
\newblock In {\em Proceedings of the IEEE/CVF International Conference on
  Computer Vision}, pages 5773--5783, 2021.

\bibitem{mildenhall2022nerf}
Ben Mildenhall, Peter Hedman, Ricardo Martin-Brualla, Pratul~P Srinivasan, and
  Jonathan~T Barron.
\newblock Nerf in the dark: High dynamic range view synthesis from noisy raw
  images.
\newblock In {\em Proceedings of the IEEE/CVF Conference on Computer Vision and
  Pattern Recognition}, pages 16190--16199, 2022.

\bibitem{huang2022hdr}
Xin Huang, Qi~Zhang, Ying Feng, Hongdong Li, Xuan Wang, and Qing Wang.
\newblock Hdr-nerf: High dynamic range neural radiance fields.
\newblock In {\em Proceedings of the IEEE/CVF Conference on Computer Vision and
  Pattern Recognition}, pages 18398--18408, 2022.

\bibitem{jun2022hdr}
Kim Jun-Seong, Kim Yu-Ji, Moon Ye-Bin, and Tae-Hyun Oh.
\newblock Hdr-plenoxels: Self-calibrating high dynamic range radiance fields.
\newblock In {\em Computer Vision--ECCV 2022: 17th European Conference, Tel
  Aviv, Israel, October 23--27, 2022, Proceedings, Part XXXII}, pages 384--401.
  Springer, 2022.

\bibitem{zhu2022nice}
Zihan Zhu, Songyou Peng, Viktor Larsson, Weiwei Xu, Hujun Bao, Zhaopeng Cui,
  Martin~R Oswald, and Marc Pollefeys.
\newblock Nice-slam: Neural implicit scalable encoding for slam.
\newblock In {\em Proceedings of the IEEE/CVF Conference on Computer Vision and
  Pattern Recognition}, pages 12786--12796, 2022.

\bibitem{yang2022vox}
Xingrui Yang, Hai Li, Hongjia Zhai, Yuhang Ming, Yuqian Liu, and Guofeng Zhang.
\newblock Vox-fusion: Dense tracking and mapping with voxel-based neural
  implicit representation.
\newblock In {\em 2022 IEEE International Symposium on Mixed and Augmented
  Reality (ISMAR)}, pages 499--507. IEEE, 2022.

\bibitem{xiangli2021citynerf}
Yuanbo Xiangli, Linning Xu, Xingang Pan, Nanxuan Zhao, Anyi Rao, Christian
  Theobalt, Bo~Dai, and Dahua Lin.
\newblock Citynerf: Building nerf at city scale.
\newblock {\em arXiv preprint arXiv:2112.05504}, 2021.

\bibitem{turki2022mega}
Haithem Turki, Deva Ramanan, and Mahadev Satyanarayanan.
\newblock Mega-nerf: Scalable construction of large-scale nerfs for virtual
  fly-throughs.
\newblock In {\em Proceedings of the IEEE/CVF Conference on Computer Vision and
  Pattern Recognition}, pages 12922--12931, 2022.

\bibitem{MartinBrualla2020NeRFIT}
Ricardo Martin-Brualla, Noha Radwan, Mehdi S.~M. Sajjadi, Jonathan~T. Barron,
  Alexey Dosovitskiy, and Daniel Duckworth.
\newblock Nerf in the wild: Neural radiance fields for unconstrained photo
  collections.
\newblock {\em 2021 IEEE/CVF Conference on Computer Vision and Pattern
  Recognition (CVPR)}, pages 7206--7215, 2020.

\bibitem{Barron2021MipNeRFAM}
Jonathan~T. Barron, Ben Mildenhall, Matthew Tancik, Peter Hedman, Ricardo
  Martin-Brualla, and Pratul~P. Srinivasan.
\newblock Mip-nerf: A multiscale representation for anti-aliasing neural
  radiance fields.
\newblock {\em 2021 IEEE/CVF International Conference on Computer Vision
  (ICCV)}, pages 5835--5844, 2021.

\bibitem{Zhang2020NeRFAA}
Kai Zhang, Gernot Riegler, Noah Snavely, and Vladlen Koltun.
\newblock Nerf++: Analyzing and improving neural radiance fields.
\newblock {\em ArXiv}, abs/2010.07492, 2020.

\bibitem{Yu2021PlenOctreesFR}
Alex Yu, Ruilong Li, Matthew Tancik, Hao Li, Ren Ng, and Angjoo Kanazawa.
\newblock Plenoctrees for real-time rendering of neural radiance fields.
\newblock {\em 2021 IEEE/CVF International Conference on Computer Vision
  (ICCV)}, pages 5732--5741, 2021.

\bibitem{lin2021barf}
Chen-Hsuan Lin, Wei-Chiu Ma, Antonio Torralba, and Simon Lucey.
\newblock Barf: Bundle-adjusting neural radiance fields.
\newblock In {\em Proceedings of the IEEE/CVF International Conference on
  Computer Vision}, pages 5741--5751, 2021.

\bibitem{li2022nerfacc}
Ruilong Li, Matthew Tancik, and Angjoo Kanazawa.
\newblock Nerfacc: A general nerf acceleration toolbox.
\newblock {\em arXiv preprint arXiv:2210.04847}, 2022.

\bibitem{barron2022mip}
Jonathan~T Barron, Ben Mildenhall, Dor Verbin, Pratul~P Srinivasan, and Peter
  Hedman.
\newblock Mip-nerf 360: Unbounded anti-aliased neural radiance fields.
\newblock In {\em Proceedings of the IEEE/CVF Conference on Computer Vision and
  Pattern Recognition}, pages 5470--5479, 2022.

\bibitem{muller2022instant}
Thomas M{\"u}ller, Alex Evans, Christoph Schied, and Alexander Keller.
\newblock Instant neural graphics primitives with a multiresolution hash
  encoding.
\newblock {\em ACM Transactions on Graphics (ToG)}, 41(4):1--15, 2022.

\bibitem{martin2021nerf}
Ricardo Martin-Brualla, Noha Radwan, Mehdi~SM Sajjadi, Jonathan~T Barron,
  Alexey Dosovitskiy, and Daniel Duckworth.
\newblock Nerf in the wild: Neural radiance fields for unconstrained photo
  collections.
\newblock In {\em Proceedings of the IEEE/CVF Conference on Computer Vision and
  Pattern Recognition}, pages 7210--7219, 2021.

\bibitem{kingma2014adam}
Diederik~P Kingma and Jimmy Ba.
\newblock Adam: A method for stochastic optimization.
\newblock {\em arXiv preprint arXiv:1412.6980}, 2014.

\end{thebibliography}

\end{document}